%% file: main.tex
\documentclass[acmsmall]{acmart}
\AtBeginDocument{%
  }

\usepackage{enumitem}

\usepackage{subcaption}
\usepackage{csquotes}
\usepackage[utf8]{inputenc}
\usepackage[binary-units]{siunitx}
\usepackage{xfp}
\usepackage{xspace}
\usepackage{microtype}
\usepackage{wrapfig}
\usepackage[check=warning,orphans=prevent,widows=prevent]{widows-and-orphans}

\usepackage{makecell}
\usepackage{pgf}
\usepackage{soul}
\usepackage{amsmath}
\usepackage{colortbl}
\usepackage{array}
\usepackage{pifont}
\usepackage{multirow}
\usepackage{footnote}
\usepackage{tabularx} 
\usepackage{tikz}
\usepackage{float}
\usetikzlibrary{positioning,fit,arrows.meta,calc}
\usepackage{graphicx} 
\usepackage{xcolor}
\tikzstyle{tier} = [rectangle, draw=black, rounded corners, minimum width=2.5cm, minimum height=1cm, align=center]
\tikzstyle{subbox} = [rectangle, draw=gray, rounded corners, fill=gray!10, inner sep=2pt, text width=3.5cm, align=left]

\usepackage{algorithm}
\usepackage{algpseudocode}

\definecolor{HoleBlue}{HTML}{1F4D8C}

\newcommand{\pointi}{\textit{(i)}}
\newcommand{\pointii}{\textit{(ii)}}
\newcommand{\pointiii}{\textit{(iii)}}
\newcommand{\pointiv}{\textit{(iv)}}
\newcommand{\pointv}{\textit{(v)}}

\usepackage{newfloat}
\usepackage{listings}
\usepackage{pgfplots}
\usepackage{pgfplotstable}
\usepackage{etoolbox}
\DeclareCaptionStyle{ruled}{labelfont=normalfont,labelsep=colon,strut=off} 
\floatstyle{ruled}
\newfloat{listing}{tb}{lst}{}
\floatname{listing}{Listing}

\usepackage[normalem]{ulem} 
\usepackage{xcolor}

\definecolor{mygreen}{rgb}{0,0.5,0} 
\definecolor{myred}{rgb}{0.6,0,0}
\definecolor{myorange}{rgb}{1.0, 0.5, 0.0}

\usepackage[markup=none, commandnameprefix=always]{changes}

\newif\ifdiffmode
\diffmodefalse 

\makeatletter
\ifdiffmode
    \renewcommand{\chadded}[2][]{\textcolor{mygreen}{#2}}
    \renewcommand{\chdeleted}[2][]{\textcolor{myred}{\sout{#2}}}
    \renewcommand{\chreplaced}[3][]{\textcolor{myorange}{#2} 
    \textcolor{myred}{\sout{#3}}}
    \renewcommand{\chcomment}[2][]{\marginpar{\scriptsize\textcolor{blue}{#2}}}

    \newenvironment{chaddedblock}{\color{mygreen}}{}
    \newenvironment{chdeletedblock}{\color{myred}\itshape}{\vspace{1em}} 
    \newenvironment{chreplacedblock}[1]{%
        \color{myorange}#1 
        \par\medskip\color{myred}\itshape --- Replaces: --- \par 
    }{%
        \par\normalcolor
    }
\else
    \renewcommand{\chadded}[2][]{#2}
    \renewcommand{\chdeleted}[2][]{}
    \renewcommand{\chreplaced}[3][]{#2}
    \renewcommand{\chcomment}[2][]{}
    \newenvironment{chaddedblock}{}{}

\fi
\makeatother

\newif\ifcomments

\commentstrue

\ifcomments
\providecommand{\CD}[1]{\textbf{\textcolor{red}{CD: #1}}}
\providecommand{\AS}[1]{\textbf{\textcolor{blue}{AS: #1}}}
\providecommand{\HP}[1]{\textbf{\textcolor{teal}{HP: #1}}}
\else
\providecommand{\CD}[1]{}
\providecommand{\AS}[1]{}
\providecommand{\HP}[1]{}
\fi

\input{macros}

\setcopyright{cc}
\setcctype{by}
\acmDOI{10.1145/3808347}
\acmYear{2026}
\acmJournal{PACMPL}
\acmVolume{10}
\acmNumber{PLDI}
\acmArticle{269}
\acmMonth{6}
\acmSubmissionID{pldi26main-p906-p}
\received{2025-11-14}
\received[accepted]{2026-04-03}

\begin{document}



\title{TreeCoder: Systematic Exploration and Optimisation of Decoding and Constraints for LLM Code Generation}

\author{Henrijs Princis}
\orcid{0009-0001-3995-0965}
\affiliation{
  \institution{University of Bristol}
  \country{United Kingdom}
}
\email{de25328@bristol.ac.uk}

\author{Arindam Sharma}
\orcid{0000-0001-5361-1057}
\affiliation{
  \institution{University of Bristol}
  \country{United Kingdom}
}
\email{arindam.sharma@bristol.ac.uk}

\author{Cristina David}
\orcid{0000-0002-9106-934X}
\affiliation{
  \institution{University of Bristol}
  \country{United Kingdom}
}
\email{cristina.david@bristol.ac.uk}








\renewcommand{\shortauthors}{Princis et al.}


\begin{abstract}
Large language models (LLMs) have shown remarkable ability to generate code, yet their outputs often violate syntactic or semantic constraints when guided only through natural language prompts. We introduce \treecoder, the most general and flexible framework to date for exploring decoding strategies, constraints, and hyperparameters in LLMs, and use it in code generation to enforce correctness and structure \emph{during decoding} rather than relying on prompt engineering.
\treecoder represents decoding as a tree search over candidate programs, where both decoding strategies and constraint functions--such as style, syntax, execution--are treated as first-class, optimisable components. This design enables systematic exploration and automatic tuning of decoding configurations using standard optimisation techniques. Experiments on Python, SQL and Rust show that \treecoder consistently improves accuracy across open-source models such as CodeLlama, Mistral, DeepSeek and Qwen, \chadded{often significantly outperforming their unconstrained baselines.}
\end{abstract}

\ccsdesc[500]{Software and its engineering~Constraints}
\ccsdesc[300]{Computing methodologies~Artificial intelligence}



\keywords{Constrained Decoding, Code Generation, Large Language Models}


\maketitle

\section{Introduction}
\label{sec:intro}

\input{sections/introduction1}

\section{Background}
\label{sec:background}
\input{sections/background}

\section{Framework for Constrained Decoding}
\label{sec:framework}
\input{sections/framework}

\section{Examples of Decoding Functions}
\label{sec:transition_functions}
\input{sections/transition_functions}

\section{Examples of Constraint Functions}
\label{sec:constraint_functions}
\input{sections/constraint_functions}


\section{Experimental Setup}
\label{sec:methodology}
\input{sections/methodology}

\section{Experimental Evaluation}
\label{sec:evaluation}

\input{sections/evaluation}


\section{Related Work}
\label{sec:related work}
\input{sections/related}

\section{Conclusions}
\label{sec:conclusions}
\input{sections/conclusions}

\section*{Data Availability Statement}
The source code for \treecoder, along with the benchmark datasets and scripts necessary to reproduce the evaluation in Section \ref{sec:evaluation}, are archived on Zenodo \cite{treecoder_code}. The source code is released under the 
Creative Commons Attribution 4.0 International license. 


\begin{acks}
Cristina David was supported by the Royal Society University Research Fellowship URF\textbackslash R\textbackslash221031. Arindam Sharma was supported by the Royal Society RF\textbackslash ERE\textbackslash221081 and the Research Institute on Verified Trustworthy Software Systems (VeTSS) through the VeTSS Research Awards (2025).
\end{acks}



\bibliographystyle{ACM-Reference-Format}
\bibliography{references}


\appendix
\input{sections/appendix}

\end{document}

%% file: macros.tex
\newcommand{\eg}{e.g.,~}
\newcommand{\ie}{i.e.~}

\newcommand{\etal}{et al.~}

\newcommand{\treecoder}{\textsc{TreeCoder}\xspace}

\newcommand{\codellama}{\textsc{CodeLlama-7B}\xspace}
\newcommand{\codellamaThirteen}{\textsc{CodeLlama-13B}\xspace}

\newcommand{\mistral}{\textsc{Mistral-7B-Instruct-v0.3}\xspace}
\newcommand{\deepseek}{\textsc{deepseek-coder-7b-instruct-v1.5}\xspace}
\newcommand{\deepseekROne}{\textsc{DeepSeek-R1-7B}\xspace}
\newcommand{\mistralshort}{\textsc{Mistral-7B}\xspace}
\newcommand{\deepseekshort}{\textsc{DeepSeek-7B}\xspace}

\newcommand{\noexp}{\textsc{No-Explanations}\xspace}
\newcommand{\ut}{\textsc{Unit-Tests}\xspace}
\newcommand{\exe}{\textsc{Executes}\xspace}
\newcommand{\comments}{\textsc{No-Comments}\xspace}
\newcommand{\syntax}{\textsc{Syntax}\xspace}
\newcommand{\type}{\textsc{Types-and-Syntax}\xspace}
\newcommand{\sig}{\textsc{Correct-Signature}\xspace}

\newcommand{\alias}{\textsc{Alias}\xspace}

\AtBeginEnvironment{algorithm}{\scriptsize} 
\AtBeginEnvironment{algorithmic}{\scriptsize} 
\AtBeginEnvironment{table}{\scriptsize}

%% file: sections/introduction1.tex
Large language models (LLMs) have become the de facto standard for a wide range of code-related tasks, including code completion, summarisation and translation~\cite{copilotReview, Summarisation, chen2025systematicliteraturereviewneural}. However, their outputs are often inconsistent, violating hard constraints such as the syntax of the target programming language~\cite{scholak-etal-2021-picard} or introducing logical errors that break compilation or execution.

A straightforward way to mitigate these issues is to improve the base model, scaling up its parameters, extending training, or curating higher-quality data. However, these approaches are computationally expensive, slow to iterate and do not provide any guarantees. A more efficient alternative is to enhance model behaviour at inference time, through methods that adjust or constrain the decoding process. Such techniques fall broadly into two categories: (1) \emph{Decoding strategies}, which adjust how the model explores its probability distribution during generation; and (2) \emph{Constrained decoding}, which restricts outputs to those that satisfy external rules such as grammatical, syntactic or logical constraints.

\paragraph{\textbf{Decoding strategies.}} LLMs generate code token by token, a process known as \emph{decoding}, by defining a conditional probability distribution \( P_\theta(x_t \mid x_{<t}) \) over the next token \(x_t\) given the preceding sequence \(x_{<t}\).

A wide range of \emph{decoding strategies} have been proposed to guide how language models (LMs) explore their predictive distribution during generation, each offering a different trade-off between efficiency, diversity, and correctness. The simplest approach, \emph{sampling}, draws tokens stochastically from the model’s conditional distribution to produce diverse outputs \cite{alphaCode, chen2021evaluatinglargelanguagemodels, zhu2023hotcoldadaptivetemperature}. More structured alternatives such as \emph{beam search} maintain the top-$k$ most likely continuations at each step \cite{meister2021beamsearchanswerquestion, huang2018finishoptimalbeamsearch,
he-etal-2023-empirical, hokamp-2017, scholak-etal-2021-picard,
lu-etal-2021-neurologic}. Closely related greedy methods such as \emph{best-first search} also prioritise high-likelihood continuations, but revisit earlier prefixes through backtracking when later validation rejects a selected branch.
Recent work has adapted more exploratory planning techniques, notably
\emph{Monte Carlo Tree Search} (MCTS) and \emph{Sequential Monte Carlo} (SMC), from games and Bayesian inference, showing promising results on mathematical reasoning and code synthesis~\cite{puri2025rolloutrouletteprobabilisticinference,
loula2025syntacticsemanticcontrollarge}. 
Each strategy embodies a distinct bias-variance compromise: sampling promotes diversity but may produce invalid code, greedy MAP-oriented methods such as beam search and best-first search prioritise likelihood at the cost of diversity, and population-based methods such as MCTS and SMC aim to balance exploration and exploitation under a fixed computational budget.
No single method is uniformly optimal across all domains or decoding budgets. 

While these decoding strategies determine \emph{how} the model explores its distribution, they do not restrict \emph{what} the model is allowed to produce. As a result, even high-likelihood sequences may violate the syntactic or semantic rules of the target language.

\paragraph{\textbf{Constrained decoding.}}
Contrary to this, \emph{constrained decoding}~\cite{geng2024grammarconstraineddecodingstructurednlp, scholak2021picardparsingincrementallyconstrained, poesia2022synchromeshreliablecodegeneration,scholak-etal-2021-picard,park2024grammaraligneddecoding,loula2025syntacticsemanticcontrollarge} modifies the decoding process by introducing a predicate \(\Phi: \Sigma^* \rightarrow \{0,1\}\) that specifies whether a partial sequence is admissible according to syntactic, semantic, or domain-specific rules. 
Formally, the constrained next-token distribution is
$P^{\Phi}_{\theta}(x_t \mid x_{<t})
    \;\propto\;
    P_{\theta}(x_t \mid x_{<t}) \cdot \Phi(x_{1:t})$.
Only continuations satisfying \(\Phi(x_{1:t}) = 1\) are considered during generation. 
As such, while decoding strategies control the search process itself, constraint-based approaches act orthogonally, enforcing external rules that restrict admissible outputs.
For instance, \emph{Adaptive Sampling with Approximate Expected Futures} (ASAp) used in Park~\etal~\cite{park2024grammaraligneddecoding}, is a \emph{sampling} strategy that samples from a LM while ensuring outputs satisfy a \emph{grammar constraint} and faithfully model the LM's conditional distribution over syntactically valid strings.

These methods are particularly valuable in \emph{code generation}, where correctness is discrete and non-negotiable: any syntactic or logical violation immediately results in compilation or runtime failure.

\paragraph{\textbf{Main challenges for constrained decoding.}}
Despite rapid progress in decoding algorithms and constraint mechanisms, research on code generation remains fragmented.
Each decoding strategy--sampling, best-first search, beam search, SMC, MCTS, ASAp--implements its own control loop, state representation and bookkeeping. Similarly, each constraint type tends to be integrated in an ad hoc manner, often coupled tightly to a specific decoding routine.

As decoding methods become increasingly specialised, researchers face two main challenges.
First, \emph{\textbf{existing systems~\cite{geng2024grammarconstraineddecodingstructurednlp,ahmed2024controllable,loula-etal-2025-smc,gad-nips-2025,geng2024sketch} hard-code particular combinations of search algorithms and constraint mechanisms}}, making it difficult to isolate their individual contributions, compare their effects, or reuse and extend implementations.
Second, \emph{\textbf{there is no principled way to systematically explore or optimise this growing space of alternatives}}.
Decoding strategies, constraints, and their hyperparameters interact in complex, non-linear ways, and small changes in the way they are chosen can dramatically alter performance.
As a result, many observed improvements are discovered through trial and error rather than through a structured, data-driven search over the design space.

The most well-known framework for training and inference with neural networks is Hugging Face \cite{wolf2020huggingfacestransformersstateoftheartnatural}. 
While dominant for model training and inference, Hugging Face only supports beam search and sampling out of the decoding strategies we study, as it does not track a decoding tree, limiting support for backtracking methods such as best-first search and for broader tree-based decoding methods.
IterGen~\cite{ugare2025itergeniterativesemanticawarestructured} addresses this partially by introducing a decoding tree and grammar-symbol navigation. However, it decodes a single hypothesis at a time and tightly couples constraints with decoding, making it difficult to flexibly combine or systematically compare alternative algorithms and constraint configurations.

\begin{chaddedblock}
\paragraph{\textbf{This work and contributions.}}
We introduce \treecoder, a unified and extensible framework for exploring decoding strategies,
constraints and hyperparameters in LLM-based code generation.

\begin{itemize}[noitemsep,topsep=0pt,leftmargin=*]
\item \textbf{\textit{TreeCoder's main theoretical contribution is a unifying abstraction for decoding: trees of partial programs and expand-update-prune-move operators acting on those trees.}} We show that our abstraction generalises many state-of-the-art decoding strategies, both population- and trajectory-based. It transforms algorithm-specific implementations into composable variations of a common structure, enabling systematic comparison and optimisation.

\item \textbf{\textit{TreeCoder's main technical contribution is the design and implementation of a modular framework built on this unified abstraction.}} It decouples decoding strategies, constraints, termination conditions, aggregation functions and optimisation into explicitly composable and comparable components. On the MBPP dataset~\cite{MBPP_dataset}, our implementation has little to no measurable overhead relative to bespoke HuggingFace implementations. Unlike prior systems like IterGen~\cite{ugare2025itergeniterativesemanticawarestructured}, \treecoder supports parallel hypotheses tracking and constraint-aware decoding in a single, extensible design.

\item \textbf{\textit{Empirically, one advantage of TreeCoder is that it can improve the performance of (smaller) open-weights models}}, 
which are particularly useful in scenarios where privacy or cost constraints favour local deployment.
\end{itemize}
\end{chaddedblock}

\paragraph{\textbf{Use of \treecoder in code generation}} 
Most current interactions with LLMs for coding rely on ``vibe coding'', steering the model through carefully worded prompts rather than explicit control.
Developers embed requirements such as ``only output valid Python code'' or ``use this API'' in natural language, but these instructions are interpreted heuristically and often ignored, especially when they conflict with the model's learned biases or extend across multiple generation steps.


For \emph{users seeking more control}, \treecoder\ replaces this trial-and-error process with a principled alternative: applying constraints \emph{during decoding} rather than merely describing them in the prompt.
In this paradigm, stylistic, syntactic and semantic rules become first-class, programmable components that dynamically steer the model's generation process, an approach particularly well-suited to code generation, which must satisfy strict syntactic and semantic correctness. \treecoder offers a comprehensive library of such constraints and users can define new ones,
making it easy to prototype, combine and experiment with new decoding behaviours.

Beyond constraints, \treecoder also allows users to interchange decoding strategies and, when datasets are available, to automatically optimise configurations of decoding strategies, constraints and hyperparameters, finding the best setting for a given coding task.



\paragraph{\textbf{Experimental results}}
We evaluated the impact of \treecoder{} on code generation across small open models (\codellama \& \codellamaThirteen~\cite{codellama}, \mistral~\cite{jiang2023mistral7b}, \deepseek~\cite{deepseekai2025deepseekv3technicalreport}, \deepseekROne~\cite{Guo_2025} and Qwen2.5-Coder-1.5B~\cite{yang2025qwen3technicalreport}). We considered generation tasks for
Python (MBPP~\cite{MBPP_dataset} and LiveCodeBench~\cite{DBLP:conf/iclr/JainHGLYZWSSS25}), SQL (SQLSpider~\cite{yu2019spiderlargescalehumanlabeleddataset}) and Rust (translated MBPP~\cite{orlanski_bc_mbpp}). Across all models and tasks, \treecoder{} yields consistent improvements in accuracy. Moreover, through automatic optimisation of decoding strategies, constraint sets and hyperparameters, \treecoder{} discovers task-specific configurations that further enhance performance.

%% file: sections/background.tex




In this section, we introduce some background concepts and terminology used throughout the paper.
For this purpose, Table~\ref{tab:decoding_functions_categorization} presents a taxonomy of decoding functions, classifying representative algorithms along five orthogonal dimensions that characterise most existing decoding strategies.

\pointi{} \emph{\textbf{Determinism}}-whether the search procedure is stochastic (\eg sampling) or deterministic (\eg beam search or best-first search).
\pointii{} \emph{\textbf{Concurrency}}-whether multiple hypotheses are expanded in parallel (\emph{population-based}) or a single trajectory is explored at a time (\emph{trajectory-based}).
\pointiii{} \emph{\textbf{Backtracking}}-whether the algorithm can revisit earlier states to explore unvisited branches.
\pointiv{} \emph{\textbf{Roll-outs}}-whether candidate continuations are recursively expanded to full completions before being evaluated; in other words, a \emph{roll-out} simulates entire trajectories from a partial hypothesis to assess its long-term promise.
\pointv{} \emph{\textbf{Optimisation goal}}-whether the algorithm seeks a single most probable completion (\emph{MAP-oriented}) or approximates the underlying model distribution (\emph{distribution-oriented}).

\emph{MAP-oriented methods}-such as beam search, best-first search, and MCTS-prioritise the most likely program under model and constraint scores.
\emph{Distribution-approximating methods}-such as sampling, SMC, and ASAp, aim to capture diverse plausible continuations, approximating the constrained posterior rather than a single mode.
Roll-out–based algorithms like MCTS and ASAp both backtrack and expand candidates to completion, while methods like best-first search perform backtracking without full roll-outs.



\begin{table}[htbp]
\centering
\begin{tabular}{l l l l l l }
\hline
\makecell{\textbf{Decoding Function}} & \makecell{\textbf{Determinism}} & \makecell{\textbf{Concurrency}} & \textbf{Backtracking} & \textbf{Roll-outs} & \makecell{\textbf{Goal Type}} \\ \hline
Beam Search                & Deterministic & Population                & No  & No  & MAP \\ \hline
SMC                        & Stochastic    & Population                & No  & No  & Distribution \\ \hline
Sampling                   & Stochastic    & Population                & No  & No  & Distribution \\ \hline
MCTS                       & Stochastic    & Trajectory                & Yes & Yes & MAP \\ \hline
ASAp                       & Stochastic    & Trajectory                & Yes & Yes & Distribution \\ \hline
Best-First Search& Deterministic & Trajectory                & Yes & No  & MAP \\ \hline
\end{tabular}
\caption{
Categorisation of decoding functions. 
}
\label{tab:decoding_functions_categorization}
\end{table}


\emph{\textbf{Instantiation within our framework.}}
To demonstrate the generality of our framework, we instantiate six representative decoding techniques: \chadded{beam search, SMC, sampling, MCTS, ASAp and best-first search.}  
Each is implemented using shared modular components for search policy and constraint enforcement, showing that diverse decoding algorithms and constraint combinations can be expressed within a single tree search abstraction.

%% file: sections/framework.tex

\begin{figure}[h]
\centering
\begin{minipage}{0.45\textwidth}
  \includegraphics[width=\linewidth]{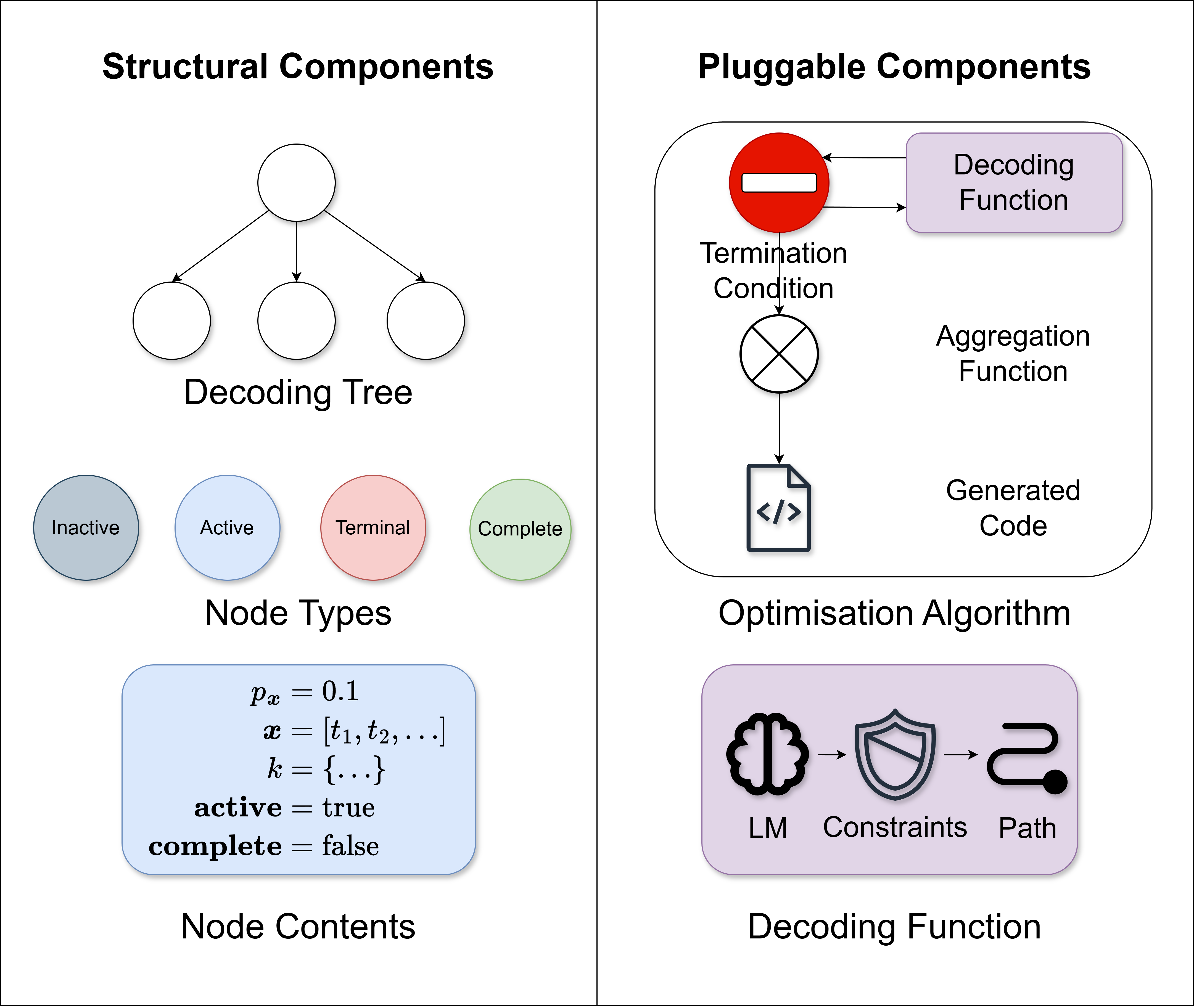}
\end{minipage}%
\hfill
\begin{minipage}{0.5\textwidth}
  \captionof{figure}{High-level illustration of our framework. 
  Structural components including the decoding tree, nodes, and node contents are on the left. 
  Pluggable components including decoding function, termination condition, aggregation function and optimisation algorithm are on the right. 
  The decoding function consists of a language model and an optional set of constraints.}
  \Description{High-level illustration of the framework showing structural and pluggable components.}
  \label{fig:framework}
\end{minipage}
\end{figure}

At a high level, the framework treats constrained code generation as a form of tree search that is guided jointly by the LLM and user-specified constraints (Figure~\ref{fig:framework}). 
Structural components, such as the decoding tree, its nodes, and their internal states, define the shared substrate over which all algorithms operate. 
Pluggable components such as the decoding function, termination condition, aggregation function, constraint modules and optimisation algorithm encapsulate the policy, stopping criteria, and external rules that govern traversal of the tree. 
By separating these structural and pluggable elements, the framework allows users to compose and compare decoding strategies systematically, to optimise configurations automatically (\eg via Bayesian optimisation), and to extend existing implementations with minimal code. 

We next provide a high-level description of the framework's components. 
We begin with the structural elements that underpin all decoding algorithms, followed by the pluggable components that control and constrain the generation process.


\paragraph*{\textbf{Structural Components}}
\begin{itemize}[noitemsep, topsep=0pt, parsep=0pt, partopsep=0pt, leftmargin=*]
    \item A decoding \textbf{tree} is the \textit{unifying abstraction} across our algorithms that minimises code duplication. Every algorithm makes use of the root node, which determines where the LM's generation starts as well as methods for accessing active, inactive, terminal and complete nodes.
    
    \item To enable \textit{systematic exploration of the solution space}, we require that every \textbf{node} contains links to parent and child nodes and that they contain a score $s$ that indicates their quality. The score is usually equal to the node's probability $p_{\boldsymbol{x}}$, but more generally \chadded{it is} calculated from the state of the tree. For example, for MCTS the score is based on number of times the node has been visited and \chreplaced{led}{lead} to a solution that did not violate constraints. 
    
    \item To enable \textit{grammar symbol navigation} (\ie the ability to track and reason about the current position within a program's grammar as tokens are generated), upon every application of a constraint function such as an incremental parser, our framework automatically stores the constraint function's state within our node's state $k$. This state can then be used by the decoding function to make decisions such as how far to backtrack or to decide which branch of the search tree to investigate next.

    
\end{itemize}

We categorise each node as being either terminal, complete, active or inactive. We expand all active nodes concurrently which enables our framework to \textit{simultaneously explore multiple hypotheses}. Furthermore, this categorisation serves as an interface between the decoding function and the termination condition. 

\begin{itemize}[noitemsep, topsep=0pt, parsep=0pt, partopsep=0pt, leftmargin=*]
    \item \textbf{Terminal nodes} have a score of zero and cannot be expanded. They represent a point where a constraint violation has been detected.
    \item \textbf{Complete nodes} contain the end-of-sequence token and have a score greater than zero. They also cannot be expanded.
    \item \textbf{Active nodes} are non-terminal nodes that will be expanded in the next iteration of the decoding function.
    \item \textbf{Inactive nodes} do not contain an end-of-sequence token and usually are not expanded by the subsequent application of the decoding function.
\end{itemize}

\paragraph*{\textbf{Pluggable Components}}
In our framework, the constraint functions, decoding functions, termination conditions, aggregation functions, LM and the optimisation algorithm denote modular components that can be plugged in. We provide a library of implementations for the first four, and users are free to experiment with custom versions as well as with different LMs and optimisation algorithms. 

\begin{itemize}[noitemsep, topsep=0pt, parsep=0pt, partopsep=0pt, leftmargin=*]
    \item An \textbf{optimisation algorithm} that searches over the rest of the pluggable components and optimises a user-provided objective function.
    \item A possibly empty set of \textbf{constraint functions }\textbf{$\Phi$} that guide the LM towards generating good quality code either by reweighting candidate continuations or by validating selected continuations lazily.
    \item A \textbf{decoding function} $\delta$ that iteratively grows the tree by expanding the active nodes using the LM and, depending on the algorithm, either eagerly or lazily applying the constraints until a terminal condition is reached.
    \item A \textbf{termination condition} $\rho$ that checks whether the tree is complete.
    \item An \textbf{aggregation function} $\otimes$ that aggregates nodes within the tree to produce the final output.
    \item An auto-regressive \textbf{LM} that given a sequence of tokens defines the probability distribution over a single token continuation. 
\end{itemize}
The pseudocode for our approach is given in Algorithm \ref{alg:Constrained Code Generation} and defines the core loop of our framework, which treats decoding as a tree search guided by a LM and constraint functions. Starting from a root node, the framework repeatedly expands active nodes using a decoding function, $\delta$, until a termination condition, $\rho(Tree)$, is met. Depending on the decoding strategy, constraints can either be incorporated eagerly into candidate scoring or evaluated lazily only on the continuation selected for expansion, as in best-first search. Once the tree is complete, an aggregation function, $\operatorname{\otimes(\textit{Tree})}$, selects the best set of candidates. 

\begin{algorithm}
\begin{algorithmic}[1]
\Require $\delta$: Decoding function  
\Require $\Phi$: Optional set of constraint functions  
\Require $\rho$: Terminal condition  
\Require $\otimes$: Aggregation function  
\Require $\operatorname{LM}$: A trained language model

\State $n \gets Node(\boldsymbol{x}=[\ ], p_{\boldsymbol{x}}=1, score=1)$ \Comment{The starting node contains an empty sequence of tokens, probability of 1, and a score of 1}
\State $Tree \gets create\_tree(n)$ \Comment{Tree is created with root node}
\While{$\rho(Tree) \neq true$} \Comment{While tree is incomplete} 
\State $Tree \gets \delta(Tree, \operatorname{LM}, \Phi)$ \Comment{Extend it by using the decoding function}
\EndWhile
\State $\textbf{return} \operatorname{\otimes(\textit{Tree})}$ \Comment{Once tree is complete, return the most promising sequences by aggregating the tree}
\end{algorithmic}
\caption{\treecoder}
\label{alg:Constrained Code Generation}
\end{algorithm}


In the rest of the section, we provide details about the individual components of the framework.
\subsection{\textbf{Optimisation Algorithm}}
\label{sec:optimisation}




\treecoder includes a higher-level optimisation module that treats decoding functions, constraint sets and hyperparameters as elements of a joint search space and applies standard optimisation methods such as grid search, random search, or Bayesian optimisation to identify effective configurations.
By abstracting optimisation as an independent module, \treecoder cleanly separates search logic from decoding, enabling systematic and reusable tuning across tasks.

Algorithm~\ref{alg:Optimisation Algorithm} outlines our implementation.
At each iteration, the optimiser proposes a new parameter configuration based on past evaluations, executes \treecoder with these parameters, and records the resulting performance.
After a predefined number of iterations, the best-performing configuration is returned.
This design cleanly separates optimisation logic from decoding and provides a minimal interface, enabling seamless integration with libraries such as Optuna~\cite{akiba2019optuna}.


\begin{algorithm}
\begin{algorithmic}[1]
\Require $\theta$: The ranges of hyper-parameters to optimise (e.g., decoding functions or population size range)
\Require $\text{optimise}$: A function that proposes new parameter instantiations based on past evaluations
\Require $max\_iters$: The maximum number of optimisation iterations
\State $past\_accuracies \gets \emptyset$
\State $past\_params \gets \emptyset$
\While{$i < max\_iters$}
    \State $params \gets \text{optimise}(\theta, past\_accuracies, past\_params)$
    \State $accuracy \gets \text{TreeCoder}(params)$
    \State $past\_accuracies \cup \{accuracy\}$
    \State $past\_params \cup \{params\}$
    \State $i += 1$
\EndWhile
\State \Return $\arg\max_{p \in past\_params} \text{value}(p)$\Comment{Return the best instantiation of the parameters the optimisation algorithm was able to find}
\end{algorithmic}
\caption{Optimisation Algorithm}
\label{alg:Optimisation Algorithm}
\end{algorithm}

\subsection{\textbf{Decoding Function $\delta$}}
\label{sec:Decoding Function}

The decoding function is a pluggable component that iteratively expands the decoding tree until a termination condition is met, as outlined in Algorithm~\ref{alg:Constrained Code Generation}. Its implementation follows four core stages: \emph{expand}, which grows active nodes using the LM’s probability distribution or performs roll-outs; \emph{update}, which revises node scores based on the optimisation goal (MAP or distribution approximation); \emph{prune}, which removes low-scoring branches and manages memory by clearing cached states; and \emph{move}, which selects the next active nodes, determining whether the algorithm performs backtracking and whether it operates in a population- or trajectory-based manner. This unified structure allows diverse algorithms to share a common control loop (Algorithm~\ref{alg:Decoding Function}) while varying in their specific implementations of these stages. We next explain each stage in more detail.

\begin{algorithm}[h]
\caption{$\delta -$ General decoding function}
\begin{algorithmic}[1]
\Require $Tree$: The current search tree  
\Require LM: Language model
\Require $\Phi$: Possibly empty set of constraints
\State $children \gets \text{Expand}(N,\text{LM},\Phi) \text{ or Rollout}(N, \text{LM}, \Phi, \delta)$\Comment{Expand step}
\State $children \gets \text{LocalUpdate}(children)$ \Comment{Update children's scores}
\State $Tree \gets \text{GlobalUpdate}(Tree)$ \Comment{Update non-child node states}
\State $Tree \gets \text{Prune}(Tree)$ \Comment{Prune non-active nodes}
\State $Tree \gets \text{Move}(Tree)$ \Comment {Select new set of active nodes}
\State $\textbf{return } Tree$
\end{algorithmic}
\label{alg:Decoding Function}
\end{algorithm}

\paragraph*{\textbf{Expand}} 

The expand step (line 1 in Algorithm~\ref{alg:Decoding Function}) either expands the set of active nodes or performs a \textit{roll-out} which is a recursive application of the general decoding function. 

Let $N$ be a set of nodes to be expanded and $n \in N$ denote a single node. The Expand operator applies the expansion procedure to each node in $N$ and returns the union of their individual expansions (see Equation~\ref{eq:expand_set}). 

We denote the set of $n$'s children as $\text{Expand}(n,\text{LM},\Phi)$. Each child node is defined by a token sequence $\boldsymbol{x} = n_{\boldsymbol{x}} \circ t$, where $\circ$ denotes concatenation and $t \in \mathcal{V}$ is a token from the LM's vocabulary (see Equation~\ref{eq:expand}). 

The child’s probability is computed as $p_{\boldsymbol{x}} = p_{\text{parent}} \cdot p(t \mid n_{\boldsymbol{x}}) \cdot \Phi(n_{\boldsymbol{x}} \circ t)$ divided by a normalising constant that ensures the probabilities of all children sum to $p_{\text{parent}}$. \begin{chaddedblock}
    In the case that no continuation is valid according to the set of constraint functions, we set all child probabilities to zero without normalizing to avoid division by zero. \end{chaddedblock}  Here $p_{\text{parent}}$ is the probability of the parent node that's being expanded, $p(t \mid n_{\boldsymbol{x}})$ is the conditional probability of choosing token $t$ and $\Phi$ is the aggregate score given by a set of constraint functions (see Equation~\ref{eq:expand_prob}).

Finally, to calculate the product of experts score given by the set of constraints $\Phi$, we take their product (see Equation~\ref{eq:phi_product}).

\noindent
\begin{minipage}[t]{0.48\columnwidth}
\small
\begin{equation}
\text{Expand}(N,\text{LM},\Phi)
= \bigcup_{n \in N} \text{Expand}(n,\text{LM},\Phi)
\label{eq:expand_set}
\end{equation}
\end{minipage}\hfill
\begin{minipage}[t]{0.48\columnwidth}
\small
\begin{equation}
\text{Expand}(n,\text{LM},\Phi) =
\left\{
\left(\boldsymbol{x} = n_{\boldsymbol{x}} \circ t,\ p_{\boldsymbol{x}} \right)
\;\middle|\; t \in \mathcal{V}
\right\}
\label{eq:expand}
\end{equation}
\end{minipage}

\vspace{0.5em}

\noindent
\begin{minipage}[t]{0.48\columnwidth}
\small
\begin{equation}
p_{\boldsymbol{x}} =
p_{\text{parent}} \cdot
\frac{
  p(t \mid n_{\boldsymbol{x}})\cdot \Phi(n_{\boldsymbol{x}} \circ t)
}{
  \sum_{t' \in \mathcal{V}}
  p(t' \mid n_{\boldsymbol{x}})\cdot \Phi(n_{\boldsymbol{x}} \circ t')
}
\label{eq:expand_prob}
\end{equation}
\end{minipage}\hfill
\begin{minipage}[t]{0.48\columnwidth}
\small
\begin{equation}
\Phi(x \circ t) =
\prod_{i=1}^{n} \phi_i(x \circ t)
\label{eq:phi_product}
\end{equation}
\end{minipage}






\paragraph{\textbf{Approximate Expand}} While equations \ref{eq:expand_set} to \ref{eq:phi_product} provide a conceptually accurate description of node expansion, in practice modern LM vocabularies are so large (32k tokens for Llama 3.1) that it is impractical to evaluate a constraint function over all possible one token continuations. Furthermore, creating a large number of nodes on each expansion would quickly run into memory limitations.

To alleviate these issues, we adapt a procedure inspired by SMC \cite{loula2025syntacticsemanticcontrollarge} where a representative subset of particles approximates the true constrained distribution. In particular, we use the approximate expansion procedure below to approximate the true-weighted distribution:

\begin{enumerate}
    \item \label{step:one} Generate the LM's distribution of single token continuations of the current node.
    \item \label{step:two} Draw $j$ samples from this distribution. 
    \item \label{step:three} Evaluate the constraint function on these samples.
\end{enumerate}

%
The generated subset in most practical circumstances provides a good approximation to the set of all nodes given by Equation \ref{eq:expand}. This is because when $j$ is the size of the LM's vocabulary $|\mathcal{V}|$ and nodes are sampled without replacement, the approximate expansion procedure is equivalent to Equation \ref{eq:expand}. 

\begin{chaddedblock}
This approximation can cause incompleteness when constraint functions are poorly aligned with the LM’s token probabilities---for example, when the prompt asks for Python code but decoding is constrained to Java. It may also be inaccurate when $j \ll |\mathcal{V}|$, since a large portion of the true probability mass may lie outside the sampled tokens.

To improve the approximation when $j$ is small, we support two sampling modes in step~\ref{step:two}: \textit{greedy}, which selects the $j$ most likely tokens according to the LM (top-$k$), and \textit{stochastic}, which samples $j$ tokens from the LM’s original distribution. Greedy sampling suits MAP-oriented decoding methods such as best-first search, beam search and MCTS, while stochastic sampling is better aligned with distribution-based methods such as constrained sampling or SMC.

We also provide an optional token allow-list that guarantees certain tokens are always considered during expansion. This is particularly useful in highly constrained domains where the LM is poorly aligned with the target language (see Section~\ref{sec:targeted-comparison}). The parameter $j$ therefore exposes an explicit trade-off between efficiency and completeness.

Finally, different tokenisations can produce the same string (e.g. \texttt{"ab"} vs.\ \texttt{"a"}+\texttt{"b"}), increasing the chance of reaching correct programs through alternative paths.
\end{chaddedblock}

\paragraph{\textbf{Roll-outs.}} For algorithms without roll-outs, all active nodes are expanded once and the step is complete. For algorithms with roll-outs, active nodes are iteratively expanded until an end condition is met (\eg generation of the EOS token). Conceptually, a roll-out can be viewed as a recursive invocation of the framework, where the inner call is restricted to a non-roll-out decoding function to prevent unbounded recursion. In our implementation, this recursion is unrolled into a loop for clarity.

\paragraph{\textbf{Update.}}
The update step updates node's scores and states as shown in lines 2-3 in Algorithm~\ref{alg:Decoding Function}. The score calculation determines the algorithm's \textit{goal}. Usually algorithms score nodes according to their probability. Sampling, beam search and best-first all fall into this category. MCTS, ASAp and SMC all use more elaborate scoring schemes. For example, MCTS uses the PUCT score \cite{2012montecarlotreesearchreview} which calculates the node's score based on the probability, number of times a roll-out \chreplaced{led}{lead} to a completion that satisfies constraints, and the number of times the node and its parent has been visited. 

There are two types of updates \textit{local updates} and \textit{global updates.} Local updates update the state of newly extended nodes. Global updates update the state of all other nodes in the tree. For example, for MCTS the global update step updates the entire lineage of rolled-out nodes.

\paragraph{\textbf{Prune.}}
\chadded{Its} purpose is memory management. For example, for beamsearch of width $k$, after the \textit{Extend loop} we will have $k^2$ possible continuations. However, only the $k$ highest scoring continuations are used, therefore we can prune the rest. 
Similarly SMC, checks if the scores of the nodes fall below a certain threshold and then conditionally resamples them. Particles that were not chosen after resampling can be pruned.

\paragraph{\textbf{Move.}}
The move step of the algorithm selects which nodes will be expanded next. Population based algorithms always move to the newly extended nodes. However, trajectory based algorithms can backtrack to earlier state. For example, the ASAp algorithm never moves away from the start node. Best-first search greedily follows the highest scoring branch of the tree, but when a selected token fails validation it backtracks to the beginning of the last line and continues from a different frontier node whose score has not been downweighted.  

\subsection{\textbf{Constraint Function $\Phi$}}
While any function that imposes constraints on code generation could be implemented directly as a decoding function, we find it useful to draw a conceptual distinction between the decoding function that does the high-level decision making for which branch to explore next and the constraint function guiding the LM towards more promising solutions. This distinction makes our framework \textit{modular} and allows swapping out constraint functions and decoding functions. It also allows different decoding functions to invoke the same constraints differently, for example by applying them to every candidate continuation or lazily only to the continuation that was selected for expansion.

To handle both soft constraints (\ie preferences that guide generation without strictly invalidating outputs) and hard constraints (\ie rules that must always be satisfied), we follow the idea of potential functions in~\cite{loula2025syntacticsemanticcontrollarge}, such that our constraint function assigns a quality score to partial or complete token sequences. This design enables a unified mechanism for constraint enforcement: the update and prune stages of the decoding function can replicate hard masking behaviour (assigning zero score to invalid continuations) or soft masking (down-weighting unlikely continuations, as in~\cite{loula2025syntacticsemanticcontrollarge}).

We further adopt the product-of-experts formulation which means that $\Phi$ can be decomposed into individual constraints $\phi$ such that $\Phi = \prod_{i=1}^{n} \phi_i$ where each $\phi_i$ takes as input the sequence of tokens $\boldsymbol{x}$ and outputs a score indicating the quality of the sequence of tokens. For example, a score of 0 would indicate that the solution is invalid. 

Product of experts formulation is desirable because it encourages breaking down the score assessment into individual modular components that can be applied individually or stacked. Such modularity helps allow for more explainable systems where the end-user can see which constraint has been violated. By using multiplication as the aggregation function any expert gets veto power over the sequence of tokens. This aligns well with the setting of constrained decoding for code generation where any syntax mistake, unit test failure, or security violation invalidates the solution. 

\paragraph*{\textbf{Increased Expressive Power}}
\label{sec: Beyond CFG Languages}
Furthermore, we find that the product of experts formulation coupled with grammar constrained decoding increases expressivity of context free languages. It allows us to recognise strings that a single context-free grammar (CFG) could not. For example, consider the following two grammars over the alphabet $\{a, b, c\}$:

{\small
\[
\begin{aligned}
G_1: \quad 
&S_1 \rightarrow A\,B \\
&A \rightarrow aA \mid \epsilon \\
&B \rightarrow bBc \mid \epsilon
\end{aligned}
\qquad
\begin{aligned}
G_2: \quad
&S_2 \rightarrow C\,D \\
&C \rightarrow aCb \mid \epsilon \\
&D \rightarrow cD \mid \epsilon
\end{aligned}
\]}
Grammar $G_1$ and $G_2$ generate the language $L_1$ and $L_2$ respectively. $L_1$ consists of any number of $a$'s followed by an equal number of $b$’s and $c$’s. Similarly $L_2$ consists of an equal number of $a$'s and $b$'s followed by any number of $c$'s. Their intersection enforces both constraints simultaneously: $L_1 = \{\, a^i b^n c^n \mid i,n \ge 0 \,\}; L_2 = \{\, a^n b^n c^j \mid n,j \ge 0 \,\}; L_1 \cap L_2 = \{\, a^n b^n c^n \mid n \ge 0 \,\}$.
%
This language, well known from formal language theory, is \emph{not} context-free~\cite{bar1961formal}. Hence, while both $L_1$ and $L_2$ are individually generated by CFGs, their intersection exceeds the expressive power of context-free grammars. Our approach, however, can represent this intersection naturally by adding two context-free grammars as decoding constraints to the expand\_node function.

\subsection{\textbf{Termination Condition $\rho$}}

This function determines whether the tree is complete. If so, it returns all terminal nodes along with their associated scores. The condition is typically based on either the number of terminal nodes, length of the solution or a set time constraint. For example, for beam search of size $k$, the termination condition $\rho$ is reached after $k$ terminal nodes have been found, i.e. $\rho \colon \left( \#\{n \mid n \text{ is terminal} \} \right) \geq k$.

\subsection{\textbf{Aggregation Function $\otimes$}}

The aggregation function combines all terminal nodes to produce an output. Two common aggregation functions are $\max$ based aggregation which chooses the top-k highest scoring nodes and $count$ based aggregation which chooses the most common node.

%% file: sections/transition_functions.tex


While enumerating all of them is impractical, our library implements three widely used algorithms---best-first search, beam search and sampling---and three emerging approaches---MCTS, SMC, and ASAp---demonstrating the generality of our framework.

To illustrate our explanation of a decoding function, we compare the implementation of different decoding functions within our framework using the four steps of \textit{expand}, \textit{update}, \textit{prune}, and \textit{move}.

\paragraph*{\textbf{Expand}}
In Figure \ref{alg:extendLoop} we see the six implementations of the expand. On the left side, we show decoding functions without roll-outs. Functions without roll-outs differ only in the way they use the Expand function. In particular, beam search and best-first search are MAP oriented methods and therefore use greedy approximation, whereas sampling and SMC are both distribution approximating methods and therefore use the stochastic approximation.

Each decoding function draws a different number of children. SMC always draws a single child node and creates duplicates in the update step if the sampled node violates constraints. Sampling adjusts for lower population size by directly drawing $\left\lceil \frac{k}{|N|} \right\rceil$ children, beam search always draws $k$ children per node resulting in $k^2$ child nodes out of which only $k$ will be kept, and best-first search also draws up to $k$ greedy candidates. 

On the right side there are two decoding functions that use rollouts. This means they go through all of the four stages of \textit{expand}, \textit{update}, \textit{prune}, \textit{move} within the \textit{expand stage}.

For MCTS this is done in the while loop (lines 4 to 11). I.e. line 6 is \textit{expand}, line 7-8 is \textit{update}, line 9 performs \textit{pruning}, and line 10 corresponds to \textit{move}. The only difference between ASAP and MCTS is in how they score nodes within the \textit{expand}.

Another important consideration when implementing a rollout based decoding function is to choose the right termination condition within the while loop. Simply relying on sampling the EOS token is not sufficient because if a mistake has been made at the beginning of a Python program, then sampling the EOS token would lead to a probability of zero being assigned by the Expand function. This could then cause the terminal node to be never selected for expansion resulting in an infinite loop. This is captured by the completeness check for MCTS (line 4) and ASAP (line 3).

\begin{figure}
    \centering
    \begin{minipage}{0.95\linewidth}
        \begin{algorithm}[H]
        \caption*{\textbf{Expand}}
        \begin{algorithmic}
            \Require $Tree$, LM, $\Phi$: Possibly empty set of constraints, $k$: Tree width constraint
        \end{algorithmic}
        \end{algorithm}
    \end{minipage}
    \vspace{0.5em}

    \begin{subfigure}[t]{0.48\linewidth}
        \begin{algorithmic}[1]
            \State $N \gets get\_active\_nodes(Tree)$
            \State $greedy \gets True$ \Comment{Greedy sampling}
            \State $j \gets k$ \Comment{j is samples per node}
            \State $M \gets \text{Expand}(N,\text{LM},\Phi,greedy,j)$
        \end{algorithmic}
        \subcaption{Beam Search}
        \vspace{1em} 

        \begin{algorithmic}[1]
            \State $N \gets get\_active\_nodes(Tree)$
            \State $greedy \gets False$\Comment{Stochastic Sampling}
            \State $j \gets \left\lceil \frac{k}{|N|} \right\rceil$
            \State $M \gets \text{Expand}(N,\text{LM},\Phi, greedy, j)$
        \end{algorithmic}
        \subcaption{Sampling}

        \begin{algorithmic}[1]
            \State $N \gets get\_active\_nodes(Tree)$
            \State $greedy \gets False$\Comment{Stochastic Sampling}
            \State $j \gets 1$
            \State $M \gets \text{Expand}(N,\text{LM},\Phi, greedy,j)$
        \end{algorithmic}
        \subcaption{SMC}

        \begin{algorithmic}[1]
            \State $N \gets get\_active\_nodes(Tree)$
            \State $greedy \gets True$ \Comment{Greedy sampling}
            \State $M \gets \text{Expand}(N,\text{LM},\{\},greedy,k)$
        \end{algorithmic}
        \subcaption{\chadded{Best-First Search}}
    \end{subfigure}
    \hfill
    \begin{subfigure}[t]{0.48\linewidth}
        \begin{algorithmic}[1]
            \State $N \gets get\_active\_nodes(Tree)$
            \State $N \gets PUCT(N)$ 
            \State $m \gets sample(N, N.scores)$
            \While{not $m$.is\_complete}
                \State $m.is\_active \gets False$
                \State $M \gets \text{Expand}(N,\text{LM},\Phi,True,k)$
                \State $M \gets PUCT(M)$
                \State $m \gets sample(M, M.scores)$
                \State $tree.prune\_kv\_cache()$
                \State $m.is\_active \gets True$
            \EndWhile
        \end{algorithmic}
        \subcaption{MCTS}

        \begin{algorithmic}[1]
            \State $N \gets get\_active\_nodes(Tree)$
            \State $m \gets N[0]$
            \While{not $m$.is\_complete}
                \State $m.is\_active \gets False$
                \State $M \gets \text{Expand}(m,\text{LM},\Phi,False,k)$
                \State $M \gets EFG(M)$\Comment{Score with efg}
                \State $m \gets sample(M, M.scores)$
                \State $tree.prune\_kv\_cache()$
                \State $m.is\_active \gets True$
            \EndWhile
        \end{algorithmic}
        \subcaption{ASAP}
    \end{subfigure}

    \caption{Expand without rollouts (left) and expand with rollouts (right). The $sample$ function returns a node from the distribution defined by normalizing the node's scores to sum to 1. The \textit{PUCT} function calculates the \textit{PUCT} score \cite{Rosin2011} and \textit{EFG} calculates the node's expected future grammaticality \cite{park2024grammaraligneddecoding}.}
    \label{alg:extendLoop}
\end{figure}

\paragraph*{\textbf{Update}}

The \textit{update} part of the algorithm performs both \textit{local} and \textit{global} updates. It is responsible for scoring nodes and updating node specific state. In Figure \ref{alg:update}, we see that the update step behaves differently for each of the decoding functions. For sampling and beam search, it scores nodes using their probability (lines 1-3). These would be considered local updates since the score calculation is only based on the newly extended nodes.

Similarly SMC only applies local updates. Node score is either 0 or 1 (line 2) depending on whether it satisfies all constraints. If the effective sample size (s) falls below a threshold (0.6), then nodes are resampled according to their scores to the population size (lines 5-8). This step ensures the population size remains consistent after an application of the decoding function.

Best-first search performs a local update by scoring nodes according to their probability (line 2). It then conditionally applies a global update based on whether the top scoring continuation satisfies constraints $\Phi$. If the constraints are not satisfied the algorithm backtracks to the beginning of the last line and multiplicatively downweights the scores along the rejected suffix so that the same invalid trajectory is less likely to be selected again (lines 3-8). This lazy validation strategy is faster while remaining faithful for hard constraints.

MCTS and ASAP both employ \textit{global updates}. They both use backpropagation (lines 2-7 and 1-8) respectively to update the entire lineage of node's scores and state. For MCTS node's state includes number of wins (\ie number of times the node \chreplaced{led}{lead} to a valid completion). For ASAp, node's state includes expected future grammaticality (EFG). Note that upon node creation, we initialise playouts and wins to be equal to zero and EFG to be equal to 1.

\begin{figure}
    \centering
    \begin{minipage}{0.95\linewidth}
       \begin{algorithm}[H]
        \caption*{\textbf{Update}}
        \end{algorithm}
    \end{minipage}
    \vspace{0.5em}

    \begin{subfigure}[t]{0.48\linewidth}
        \begin{algorithmic}[1]
            \For{$m \in M$}
                \State $m.score \gets m.probability$
            \EndFor
            \State $M \gets sorted(M,M.score)$
        \end{algorithmic}
        \subcaption{Beam Search}
        \vspace{1em} 
        
        \begin{algorithmic}[1]
            \For{$m \in M$}
                \State $m.score \gets m.probability$
            \EndFor
        \end{algorithmic}
        \subcaption{Sampling}

        \begin{algorithmic}[1]
            \For{$m \in M$}
                \State $m.score \gets m.score \times \lceil m.probability \rceil$
            \EndFor
            \State $s \gets effective\_sample\_size(Nodes)$
            \If{$s \leq 0.6$}
                \State $M.scores \gets \frac{\sum M.scores}{k}$
                \State $M \gets reasample(M, M.scores,k)$
            \EndIf
        \end{algorithmic}
        \subcaption{SMC}
    \end{subfigure}
    \hfill
    \begin{subfigure}[t]{0.48\linewidth}
        \begin{algorithmic}[1]
            \State $is\_win \gets \lceil m.probability \rceil$
            \While{$m.parent$}
                \State $m.playouts \gets m.playouts +1$
                \State $m.wins \gets m.wins + is\_win$
                \State $m.score \gets PUCT(m)$
                \State $m \gets m.parent$
            \EndWhile
        \end{algorithmic}
        \subcaption{MCTS}

        \begin{algorithmic}[1]
            \While{$m.parent$}
                \State $efg \gets 0$
                \For{$c \in m.children$}
                    \State $efg \gets efg+c.probability \times c.efg$
                \EndFor
                \State $m.efg \gets efg$
                \State $m \gets m.parent$
            \EndWhile
        \end{algorithmic}
        \subcaption{ASAP}

        \begin{algorithmic}[1]
            \State $m \gets M[0]$ \Comment{Select the highest-probability continuation}
            \State $m.score \gets m.probability$
            \If{$\Phi(m) = 0$}
                \State $n \gets backtrack\_to\_start\_line(m)$
                \State $downweight\_path(n, m)$
                \State $m.score \gets 0$
                \State $m \gets n$
            \EndIf
        \end{algorithmic}
        \subcaption{Best-First Search}
        
    \end{subfigure}

    \caption{Update step of all six decoding functions}
    \label{alg:update}
\end{figure}

\paragraph*{\textbf{Prune}}
In Figure \ref{alg:prune and move} we see that that the prune step's purpose is to manage memory. The memory footprint required for storing each node's state is relatively small when using our approximation. However, KV \chadded{(Key-Value) cache~\cite{liu2025kv}} takes up a lot of space (up to 0.5GB in this work, exact size varies based on model size, architecture, and sequence length) and thus needed to be cleared on every application of decoding function to avoid out-of-memory errors. Therefore, after each update step, we perform a call to $tree.prune\_kv\_cache()$ which clears the KV cache of all inactive nodes. To maintain fast decoding, it keeps the cache of active nodes that will be expanded within the next iteration.



\paragraph{\textbf{Move}}

The move step of the algorithm selects the next node to expand. We see that population based methods (Beam Search, Sampling and SMC) all move. I.e. they change which nodes are active after an application of the decoding function. MCTS and ASAP are both stationary meaning that they never move from the tree's root node. Best-first search is trajectory-based: it activates the highest scoring expandable node in the tree, which may lie on a different branch after a rejected continuation has been downweighted and the search has backtracked to the start of the last line. Even though MCTS and ASAP remain at the root during move, their rolled-out sequences are still stored within the tree and used to calculate the end condition $\rho$ and within the aggregation function $\otimes$ to allow the framework to terminate and return the most promising sequences.

\begin{figure}
    \centering
    \begin{minipage}{0.95\linewidth}
        \begin{algorithm}[H]
        \caption*{\textbf{Prune \& Move}}
        \end{algorithm}
    \end{minipage}
    \vspace{0.5em}

    \begin{subfigure}[t]{0.48\linewidth}
        \begin{algorithmic}[1]
            \State $M \gets M[0:k]$
            \State $tree.prune\_kv\_cache()$
            \State \textbf{return} $set\_active(M)$
        \end{algorithmic}
        \subcaption{Beam Search}
        \vspace{1em} 
        
        \begin{algorithmic}[1]
            \State $M \gets \{m | m\in M \cap m.score \geq 0\}$
            \State $tree.prune\_kv\_cache()$
            \State \textbf{return} $set\_active(M)$
        \end{algorithmic}
        \subcaption{Sampling}

        \begin{algorithmic}[1]
            \State $tree.prune\_kv\_cache()$
            \State \textbf{return} $set\_active(M)$
        \end{algorithmic}
        \subcaption{SMC}
    \end{subfigure}
    \hfill
    \begin{subfigure}[t]{0.48\linewidth}
        \begin{algorithmic}[1]
            \State $tree.prune\_kv\_cache()$
            \State \textbf{return} $set\_active(tree.root\_node)$
        \end{algorithmic}
        \subcaption{MCTS}

        \begin{algorithmic}[1]
            \State $tree.prune\_kv\_cache()$
            \State \textbf{return} $set\_active(tree.root\_node)$
        \end{algorithmic}
        \subcaption{ASAP}

        \begin{algorithmic}[1]
            \State $tree.prune\_kv\_cache()$
            \State \textbf{return} $set\_active(\{m\})$
        \end{algorithmic}
        \subcaption{Best-First Search}
    \end{subfigure}

    \caption{Prune and move steps of all six decoding functions}
    \label{alg:prune and move}
\end{figure}

%% file: sections/constraint_functions.tex
In this section, we provide examples of the types of constraint functions for our framework. For illustration, Figure~\ref{fig:constraintFunctions} provides several constraint functions that can be combined within our framework.
Each constraint function takes in a sequence of tokens generated so far, a tokenizer, prefix length and optional state that contains additional information about the problem such as unit tests.
Then, it evaluates the partially generated sequence and returns a scalar score indicating its validity.
For instance, \syntax checks whether the decoded fragment can be compiled using Python's \texttt{codeop} module,
\comments suppresses inclusion of comments,
\noexp enforces that outputs start with \texttt{def} and contain no capital letters after the first line, nudging the model toward code-like continuations rather than natural language explanations,
and \ut executes generated code against extracted assertions. 


Beyond those in Figure~\ref{fig:constraintFunctions}, we define several additional constraint types that capture increasingly stricter properties of the generated programs (for brevity, these are not provided):

(i) \sig---checks whether the function names match the specification and have the correct number of arguments.

\begin{wrapfigure}{l}{0.45\linewidth}
\centering
\begin{tikzpicture}[
  node distance=7mm and 9mm,
  every node/.style={draw=none, inner sep=1pt, font=\small,scale=0.80},
  ->, thick
]
\node (syn)  {\syntax};
\node (synt) [above=of syn] {\type};
\node (exe)  [above=of synt] {\exe};
\node (ut)   [above=of exe] {\ut};

\node (sig) [left=7mm of exe] {\sig};

\draw (syn)  -- (synt);
\draw (synt) -- (exe);
\draw (exe)  -- (ut);
\draw (sig)  -- (ut);

\path[every node/.style={font=\scriptsize, sloped, anchor=center}];
\end{tikzpicture}
\caption{Partial order of constraints.} 
\label{fig:constraint-poset}
\end{wrapfigure}

(ii) \type---applies the Pyright static analyser \cite{pyright2025} to detect type and syntax errors. For this purpose, we require the model to generate explicit type annotations. (iii) \exe---verifies whether the function executes without errors when provided the correct arguments as input. Besides static errors (e.g. syntax, types), this has the ability to catch runtime errors such as infinite loops or unhandled exceptions.

These constraints form a hierarchy of restrictiveness, with \ut\ being the strongest (while \comments\ acts orthogonally):


These constraints can act individually or jointly during decoding, guiding the search toward syntactically valid, well-structured, and semantically meaningful programs (more details on the constraints and their use are given in the experimental evaluation in Section~\ref{sec:evaluation}). 
Formally, they are composed under the product-of-experts formulation introduced earlier,
where the overall constraint score is the product of individual constraint outputs.
This formulation naturally supports modular composition:
each constraint functions as an ``expert'' that can independently veto invalid continuations,
while their combination yields a single, unified quality signal used by the decoding function.
As a result, \treecoder enables complex constraint behaviours---such as enforcing syntax, style, and functional correctness simultaneously---without altering the underlying decoding logic.

\paragraph{\textbf{Expensive constraints still guide search.}} Certain constraints such as \ut and \exe can only be applied once a sequence has been fully generated. For simple decoding strategies such as sampling, this information is completely dismissed and we are just as likely to re-produce the same erroneous solution on the next iteration. In contrast, more sophisticated strategies such as MCTS and ASAp retain this information within the search tree, reducing the likelihood of repeating previous mistakes.
This means that the harder constraints make a bigger difference for more challenging problems that the model does not get right on the first try. 




\begin{figure}
    \centering
    \begin{minipage}{0.95\linewidth}
        \begin{algorithmic}
            \Require $token\_ids$: Model-generated token sequence
            \Require $tokenizer$: Tokenizer for decoding token sequences
            \Require $prefix\_length$: Index marking end of user's prompt
            \Require $state$: Optional persistent constraint state that includes \textit{unit tests}. 
        \end{algorithmic}
    \end{minipage}
    \vspace{0.5em}

    \begin{subfigure}[t]{0.48\linewidth}
        \begin{algorithmic}[1]
            \State $\mathbf{import}\ codeop$
            \State \textbf{def} python\_syntax(...):
            \State \hspace{1em} $code \gets tokenizer.decode(token\_ids)$
            \State \hspace{1em} $code \gets code[prefix\_length:]$
            \State \hspace{1em} $code \gets trim\_last\_line(code)$
            \State \hspace{1em} $valid \gets codeop.compile(code)$
            \State \hspace{1em} $score \gets 1$ if $valid$ else 0
            \State \hspace{1em} \Return ConstraintOutput(score=score)
        \end{algorithmic}
        \subcaption{Python \syntax}
    \end{subfigure}
    \hfill
    \begin{subfigure}[t]{0.48\linewidth}
        \begin{algorithmic}[1]
            \State \textbf{def} python\_comments(...):
            \State \hspace{1em} $code \gets tokenizer.decode(token\_ids)$
            \State \hspace{1em} $code \gets code[prefix\_length:]$
            \State \hspace{1em} $valid \gets$ \textbf{not}((``\#'' $\in code$) \textbf{or} ($''' \in code$))
            \State \hspace{1em} $score \gets 1$ if $valid$ else 0
            \State \hspace{1em} \Return ConstraintOutput(score=score)
        \end{algorithmic}
        \subcaption{\comments}
    \end{subfigure}

    \vspace{0.75em}

    \begin{subfigure}[t]{0.48\linewidth}
        \begin{algorithmic}[1]
            \State \textbf{def} no\_explanantions\_constraint(...):
            \State \hspace{1em} $code \gets tokenizer.decode(token\_ids)$
            \State \hspace{1em} $has\_def \gets code.startswith(``def")$
            \State \hspace{1em} $lines \gets code.split(\text{``\textbackslash n''},1)$
            \State \hspace{1em} $no\_upper \gets \neg get\_upper($lines[1]$)$
            \State \hspace{1em} $score \gets 1$ if $has\_def\ \&\ no\_upper$ else 0
            \State \hspace{1em} \Return ConstraintOutput(score=score)
        \end{algorithmic}
        \subcaption{\noexp}
    \end{subfigure}
    \hfill
    \begin{subfigure}[t]{0.48\linewidth}
        \begin{algorithmic}[1]
            \State \textbf{import} eval
            \State \textbf{def} unit\_test(...):
            \State \hspace{1em} \textbf{if }{$\neg sequence\_finished(token\_ids)$}
                \State \hspace{2em} \Return ConstraintOutput(score=1)
            \State \hspace{1em} \textbf{end if}
            \State \hspace{1em} $code \gets tokenizer.decode(token\_ids[prefix\_length:])$
            \State \hspace{1em} $tests \gets state$
            \State \hspace{1em} $pass \gets eval(code, tests)$
            \State \hspace{1em} $score \gets 1$ if $pass$ else 0
            \State \hspace{1em} \Return ConstraintOutput(score=score)
        \end{algorithmic}
        \subcaption{\ut}
    \end{subfigure}


    \caption{Constraint function examples: \syntax (checks syntactic validity), \comments (suppress comments), \noexp (encourage code over natural language explanations), \ut (validate generated code via assertions).} 
    \label{fig:constraintFunctions}
\end{figure}

%% file: sections/methodology.tex
\paragraph{\textbf{Infrastructure.}} All experiments were performed on AWS EC2 g6.2xlarge instances equipped with 8vCPUs, \SI{32}{GB} of RAM and an L4 GPU with \SI{24}{GB} of VRAM.

\paragraph{\textbf{Datasets.}} We use MBPP~\cite{MBPP_dataset} and LiveCodeBench~\cite{DBLP:conf/iclr/JainHGLYZWSSS25} for Python, SQLSpider~\cite{yu2019spiderlargescalehumanlabeleddataset} for SQL, and translated MBPP~\cite{orlanski_bc_mbpp} for Rust. For Python and Rust, each problem includes unit tests, which we split into \textit{public} tests--appended to the problem description during generation--and \textit{private} tests, which, together with the public tests, are used to measure true accuracy.
For SQLSpider, the dataset also contains a schema and a database which is used to verify the correctness of a query by executing it and comparing the result to the reference query.

%


\paragraph{\textbf{Language Models.}} We evaluate our framework using \codellama~\cite{codellama} instruct, \codellamaThirteen~\cite{codellama} instruct, \mistral \cite{jiang2023mistral7b}, \deepseek \cite{deepseekai2025deepseekv3technicalreport}, \linebreak 
\deepseekROne~\cite{Guo_2025} and Qwen2.5-Coder-1.5B~\cite{yang2025qwen3technicalreport} models. We run models whose parameter count is above \SI{4}{B} in half precision (fp16) due to VRam constraints.

\paragraph{\textbf{Framework parameters.}} \chadded{We use the following \textbf{termination condition} $\rho$ for all of our experiments: we finish generation when we find 1 complete node (corresponding to the number of solutions we are interested in), or have reached 10\,000 nodes which is an estimate of how many we can fit into our memory, or have no more active nodes, or when one of our nodes exceeds the length limit (to prevent infinite generation). The termination condition can be adjusted by the user. Also, the user can add a soft time constraint on the maximum runtime of the algorithm (e.g. we set this to be \SI{10}{s} when finding the best LM configuration in Section~\ref{sec:systematic-exploration}).}
{\small
\[
\begin{aligned}
\rho \colon\quad \#\{n \mid n \text{ is terminal \& n's score} \geq 0 \} &\geq \chadded{1}  & or \qquad&  
\#n \geq 10\,000 \qquad or \\
\#\{n \mid n \text{ is active} \} &= 0 & or \qquad &  
\#\{n \mid \text{len}(n) \geq 256 \} \geq 1 \text{ }
\end{aligned}
\]}

We use the following \textbf{aggregation function} for all of our experiments: we preferably choose nodes who are complete and have a score that's higher than 0, followed by complete nodes that contain syntax errors:
\[
\small
\otimes \colon\quad top_5(\{n \mid n \text{ is terminal} \} \  \cup 
\{n \mid n \text{ is active \& not terminal} \}\  \cup 
\{n \mid n \text{ is not active \& not terminal} \})
\]

\paragraph{\textbf{Metrics.}} 
We report the accuracy which signifies the proportion of problems where the solution passes all public and private testcases. Conversely, the unit test constraint only uses the \textit{public} tests to guide generation.

For later comparisons with other frameworks, we expand on this metric by using the pass@k metric, which checks whether any of the k generated solutions satisfy all unit tests. 
We report the total number of node expansions as a measure of how efficient our implementation is. In the context of comparing algorithms, we use the number of node expansions as an upper bound to performance. This score directly measures how many times an LLM call was made multiplied by the batchsize to produce the probability distribution over the next token. 
%

%% file: sections/evaluation.tex
\subsection{Using \treecoder for Constraint Enforcement during Decoding} 
\label{sec:targeted-comparison}


Most current interactions with LLMs for programming rely on prompt engineering to steer model behaviour and describe desired properties such as correctness or style. However, these textual constraints are interpreted heuristically and are often ignored during generation.
Conversely, \treecoder\ enables easy experimentation with constraints directly during decoding.



To illustrate this, we evaluate \codellama under a range of constraint configurations (described in Section~\ref{sec:constraint_functions}), all capturing information already described in the prompt. 
As rows 1-7 in Table~\ref{tab:model_results} show, each constraint type progressively increases the precision of the generated code. 
 In particular, unit tests yield the largest gains, while syntactic constraints alone provide marginal benefits. Lightweight constraints such as comment suppression reduce node expansions by 6.5\% compared to no constraints, without degrading accuracy. 

\begin{table}[htbp]
\centering
\begin{tabular}{lrlllrr}
\toprule
\textbf{Model} & \textbf{Width} & \textbf{Decoding} & \textbf{Constraints} & \textbf{Accuracy (\%)} & \makecell{\textbf{Expansions}} & \makecell{\textbf{\chadded{Mean time}}\\\textbf{\chadded{taken (s)}}} \\
\midrule
\codellama & 5 & Sampling & \textsc{No-Constraints} & 32.3 & 274 & 2.31 \\
\codellama & 5 & Sampling & \comments & 32.8 & 256 & 2.22 \\
\codellama & 5 & Sampling & \syntax & 34.0 & 270 & 2.36 \\
\codellama & 5 & Sampling & \sig & 36.1 & 254 & 3.57 \\
\codellama & 5 & Sampling & \type & 47.5 & 610 & 27.06 \\
\codellama & 5 & Sampling & \exe & 47.1 & 359 & 5.07 \\
\codellama & 5 & Sampling &\ut & 60.0 & 417 & 3.90 \\
\midrule
\codellama & 5 & Sampling & All & 60.7 & 316 & 3.23 \\
\codellama & 5 & SMC & All & 61.4 & 332 & 5.47 \\
\codellama & 5 & Beam Search & All & 63.0 & 706 & 16.10 \\
\codellama & 5 & MCTS & All & 65.3 & 1099 & 78.39 \\
\codellama & 5 & ASAP & All & 68.9 & 1158 & 84.05 \\
\bottomrule
\end{tabular}
\caption{Performance under different decoding strategies and constraint combinations on MBPP. “All” combines \syntax, \ut and \comments constraints. For \type checking we use the playwright library and instruct the LM to generate type annotations (which slightly increases the length of generated program, which in turn increases the node expansions). }
\label{tab:model_results}
\end{table}




For comparison, we analyse which constraints are most often violated when specified only in the prompt, especially those related to execution and types.
We find that \ut\ is most often violated (55.6\%), followed by \exe\ (39.6\%) and \type\ (25.9\%). Others are less frequent: \noexp\ (12.9\%), \comments\ (4.1\%), \syntax\ (2.4\%), and \sig\ (1.9\%).
Overall, this shows that many natural-language constraints are ignored by the LM, highlighting the need for decoding-time enforcement over prompt-level guidance.

\paragraph{\textbf{Mis-aligned models}}
The previous experiments evaluated constraint-guided coding on well-aligned models that generally followed code-only instructions. However, many open-source models such as \mistralshort\ and \deepseekshort\ exhibit persistent misalignment: they tend to ignore prompts and interleave natural-language commentary with code, producing \textit{explanation–code–explanation} patterns \emph{even when explicitly asked to output only code}. This behaviour illustrates the limitations of prompt-level control and motivates decoding-time constraint enforcement.

 To address this, we employ the lightweight \noexp constraint (Section~\ref{sec:constraint_functions}), which enforces that outputs start with \texttt{def} and contain no capital letters after the first line. This nudges the model toward code-like continuations from the outset. Because the initial \texttt{def} token often has low model confidence, we include it in the always-considered token list and slightly increase the expansion parameter ($j{=}25$ rather than $j{=}5$) to preserve plausible continuations.

As shown in Table~\ref{tab:mistral}, these minimal decoding-time interventions substantially improve code validity. For both \mistralshort\ and \deepseekshort, enforcing \noexp improves performance from
near-zero compliance with basic output instructions (denoted as - in the table) to reliable code generation in over 30\% and 40\%
of cases, respectively. 
For \mistralshort, simple syntactic checks are insufficient. They typically trigger only after invalid prefixes denoting explanations have already been generated, making recovery difficult.
This experiment reinforces the idea that even lightweight constraints, applied during decoding rather than expressed in the prompt, can realign misbehaving models toward syntactically valid and executable code.

\begin{chaddedblock}
\paragraph{\textbf{Reasoning models.}}
The No-Explanation and Syntax constraints are also useful for reasoning models such as \deepseekROne. Although given formatting instructions in the prompt, \deepseekROne{} does not obey them. Thus, we use these constraints to enforce runnable code, applying them only after “thinking” has finished.
As shown in Table~\ref{tab:mistral}, performance improves from near-zero compliance (denoted as -) to reliable code generation in over 50\% of cases. Moreover, the constraints nudge the LM to find solutions faster.
\end{chaddedblock}

\begin{table}
\centering
\begin{tabular}{lllrrr}
\toprule
\textbf{Model} & \makecell{\textbf{Decoding}\\\textbf{Function}} & \textbf{Constraints} &  \makecell{\textbf{Accuracy (\%)}} & \makecell{\textbf{Expansions}} & \makecell{\chadded{\textbf{Mean time}}\\\textbf{\chadded{taken (s)}}} \\
\midrule
\textsc{\mistralshort} & Beam Search & \textsc{no-constraints} & - & 205 & 15.83 \\
\textsc{\mistralshort} & Beam Search & \syntax & - & 603 & 27.68 \\
\textsc{\mistralshort} & Beam Search & \noexp & 33.7 & 252 & 31.26 \\
\textsc{\mistralshort} & Beam Search & \syntax+\noexp & 33.2 & 607 & 40.99 \\
\midrule
\textsc{\deepseekshort} & Beam Search & \textsc{no-constraints} & - & 190 & 10.50 \\
\textsc{\deepseekshort} & Beam Search & \syntax & 42.0 & 240 & 16.00 \\
\textsc{\deepseekshort} & Beam Search & \noexp & 40.0 & 325 & 21.00 \\
\textsc{\deepseekshort} & Beam Search & \syntax+\noexp & 42.0 & 320 & 25.04 \\
\midrule
\textsc{\deepseekROne} & Beam Search & \textsc{no-constraints} & - & 1288 & 92.28 \\
\textsc{\deepseekROne} & Beam Search & \syntax & - & 1151 & 84.32 \\
\textsc{\deepseekROne} & Beam Search & \noexp & 54.6 & 1138 & 87.74 \\
\textsc{\deepseekROne} & Beam Search & \syntax+\noexp & 52.6 & 1121 & 74.48 \\
\bottomrule
\end{tabular}
\caption{ 
    Population size was set to 5, and the expansion approximation parameter j was set to 25 (for \mistralshort and \deepseekshort). \deepseekROne results use population size 2, j=2 to keep the runtime low. "-" denotes near-zero compliance with basic output instructions.
} 
\label{tab:mistral}
\end{table}

\subsection{Using \treecoder to Compare Decoding Strategies}
\label{sec:decoding-strategies}
Beyond constraints, \treecoder{} enables users to vary the decoding strategy, revealing how different
search behaviours trade off accuracy and efficiency. Among the methods tested, ASAp achieves the highest accuracy, closely followed by MCTS (rows 8--12 in Table~\ref{tab:model_results}). Both rely on rollouts and backtracking to refine partial hypotheses, improving solution quality but increasing cost, as reflected in the higher number of expansions. In contrast, beam search expands more nodes than sampling or SMC, as it maintains distinct beams rather than merging similar candidates. These results highlight how \treecoder{} facilitates systematic exploration of decoding dynamics, allowing practitioners to switch strategies within a unified framework.



\subsection{Using \treecoder for Systematic Exploration of the Design Space}
\label{sec:systematic-exploration}

Beyond enabling users to manually select constraints and decoding strategies, \treecoder also supports \emph{automatic exploration} of this design space. In particular, it integrates the optimisation algorithm described in Section~\ref{sec:optimisation} to systematically search for configurations that maximise task performance.

\paragraph{\textbf{Finding the best configuration for \codellama}}
To illustrate this capability, we investigate the following question: \emph{What is the the optimal decoding function, constraints, and population size that optimises the accuracy given a soft time limit of \SI{10}{seconds} per sample?} \chadded{We conduct this analysis for two models--\codellama and CodeLlama-13B--with results reported in Table ~\ref{tab:decoding_flat_public_private}.} We address this question through a two-stage experiment that automatically identifies effective decoding configurations and then evaluates their generalisation to unseen test cases (note that we split the MBPP dataset into a smaller \textit{validation} dataset used to search for optimal constraint combinations and \textit{test} dataset).

\paragraph{Stage 1 - Searching for the optimal configuration.}
We employ the Optuna library \cite{akiba2019optunanextgenerationhyperparameteroptimization}, which implements Gaussian-process-based Bayesian optimisation, to explore five decoding functions (beam search, sampling, MCTS, ASAp, and SMC), population sizes from 1 to 5, and combinations of constraints including Python \syntax, \ut and \comments.
In total, the search space comprises 200 unique configurations, and the optimisation uses log Expected Improvement as the acquisition function to balance exploration and exploitation.
Each configuration is evaluated on the 50 validation problems of MBPP, with a soft time limit of \SI{10}{seconds} per query checked after each decoding step.
Table ~\ref{tab:decoding_flat_public_private} reports the best configuration found for each decoding function showing the accuracy and sample time on the test problems.

\chadded{\emph{For \codellama, the search identifies SMC and beam search with the constraints shown in Table~\ref{tab:decoding_flat_public_private} as the most effective strategies, achieving an accuracy of \SI{66}{\percent}, while staying within the runtime budget. For CodeLlama-13B, the strategies with the highest accuracy are sampling and beam search which achieve an accuracy of \SI{70}{\percent}.}}
Across decoding methods, unit tests emerge as the most influential constraint, and larger population sizes consistently improve accuracy, highlighting the benefits of maintaining multiple concurrent hypotheses.

\chadded{Additionally, for \codellama, Figure~\ref{fig:pareto_frontier} shows the Pareto frontier of accuracy against runtime on the validation dataset, allowing practitioners to select the configuration that best fits their time budget. Constrained sampling and beam search consistently dominate the frontier, offering the best accuracy-efficiency trade-off.}

\begin{table}[htbp]
\centering
\scriptsize
\setlength{\tabcolsep}{4pt}
\begin{tabular}{@{}m{1.6cm} m{2.5cm} m{2.9cm} r r r r r@{}}
\toprule
\textbf{Model} &
\textbf{Decoding Function} &
\textbf{Constraint Config.} &
\makecell{\textbf{Pop. Size}} &
\makecell{\textbf{Validation}\\\textbf{Acc.(\%)}} &
\makecell{\textbf{Validation}\\\textbf{Time(s)}} &
\makecell{\textbf{Test}\\\textbf{Acc.(\%)}} &
\makecell{\textbf{Test}\\\textbf{Time(s)}} \\
\midrule
\codellama & Sequential Monte Carlo  & \ut+\comments         & 5 & 66.0 & 2.6 & 59.2 & 3.2 \\
\codellama & Beam Search             & \ut+\syntax+\comments & 5 & 66.0 & 6.7 & 57.8 & 7.5 \\
\codellama & Sampling                & \ut+\syntax           & 4 & 62.0 & 2.7 & 55.7 & 3.3 \\
\codellama & ASAp                    & No Constraints        & 4 & 54.0 & 5.4 & 49.9 & 6.3 \\
\codellama & Monte Carlo Tree Search & \ut+\syntax           & 5 & 54.0 & 9.0 & 50.4 & 9.2 \\
\midrule
\codellama & Baseline (Sampling)     & No Constraints        & 1 & -    & -    & 49.6 & 1.9 \\
\specialrule{1pt}{1pt}{1pt}
\chadded{CodeLlama-13B} & Sampling                & \ut+\comments & 5 & 70.0 & 4.6  & 65.0 & 5.7 \\
\chadded{CodeLlama-13B} & Beam Search             & \ut           & 4 & 70.0 & 9.9  & 61.3 & 10.9 \\
\chadded{CodeLlama-13B} & Sequential Monte Carlo  & \ut           & 5 & 68.0 & 4.9  & 67.1 & 5.7 \\
\chadded{CodeLlama-13B} & ASAp                    & \ut           & 4 & 46.0 & 19.8 & 47.5 & 19.8 \\
\chadded{CodeLlama-13B} & Monte Carlo Tree Search & No Constraints & 5 & 18.0 & 18.9 & 25.5 & 18.6 \\
\midrule
\chadded{CodeLlama-13B} & Baseline (Sampling)     & No Constraints & 1 & -    & -    & 54.1 & 2.4 \\
\bottomrule
\end{tabular}
\caption{Performance of each decoding function on validation (tuning) and test splits of the MBPP dataset. For MCTS and ASAp, population size corresponds to the expansion parameter $j$.}
\label{tab:decoding_flat_public_private}
\end{table}

\paragraph{Stage 2 – Evaluate on the Test Set.}
We then test the best configurations on the test dataset split of the MBPP to assess whether the optimised settings generalise to unseen programs.
Table ~\ref{tab:decoding_flat_public_private} also presents the corresponding results on the validation split.
While overall accuracies decrease slightly, reflecting the higher difficulty of the test set, the same qualitative trends persist: \emph{configurations that incorporate unit-test constraints and use larger populations consistently outperform the unconstrained baseline}, \ie the default implementations for \codellama~ \chadded{and CodeLlama 13B which use sampling and generate a single sequence without any constraints.}  

\begin{figure}[t]
\centering
\begin{minipage}{0.6\linewidth}
\centering
\begin{tikzpicture}
\begin{axis}[
    width=\linewidth,
    height=0.7\linewidth, 
    xlabel={Time per Sample (s)},
    ylabel={Accuracy (\%)},
    title={Accuracy vs Time per Sample},
    grid=major,
    xmin=0,
    ymin=0,
    ymax=80,
    enlargelimits=false,
   font=\footnotesize,
]

\addplot[
    only marks,
    mark=*,
    mark size=1.5pt,
    color=gray,
    opacity=0.5
] coordinates {
    (1.3876772547,52.0)
    (1.4103347588,0.0)
    (1.4633009624,40.0)
    (1.7100105095,44.0)
    (1.9257155132,46.0)
    (2.0051812410,42.0)
    (2.1592948627,44.0)
    (2.2183238554,46.0)
    (2.3859643459,60.0)
    (2.5052238989,58.0)
    (2.6314294338,58.0)
    (2.6459735918,58.0)
    (2.6604421854,60.0)
    (2.7112732744,62.0)
    (2.7708250093,66.0)
    (2.8540470552,60.0)
    (2.9232355738,56.0)
    (2.9615342236,64.0)
    (3.3302865934,56.0)
    (3.9936231518,60.0)
    (5.1161245060,58.0)
    (5.2885517454,62.0)
    (5.3698417854,62.0)
    (5.3778226376,44.0)
    (5.4367986250,54.0)
    (6.0640896320,64.0)
    (6.1034233284,64.0)
    (6.1290379095,62.0)
    (6.3208829212,62.0)
    (6.5544449472,50.0)
    (6.6386376047,64.0)
    (6.6434920073,64.0)
    (6.6487682486,62.0)
    (6.6736303520,66.0)
    (6.7278102207,64.0)
    (6.7401265860,64.0)
    (6.7593546391,64.0)
    (6.7979398489,64.0)
    (6.8462254381,48.0)
    (6.9331986618,64.0)
    (6.9459114885,62.0)
    (7.0864864206,62.0)
    (7.6653053713,52.0)
    (8.4633966637,52.0)
    (9.0066353893,54.0)
    (9.7596386290,52.0)
};

\addplot[
    thick,
    color=blue,
    mark=none
] coordinates {
    (1.3845782328,52.0)
    (1.6707399416,52.0)
    (1.6707399416,56.0)
    (2.3037416697,56.0)
    (2.3037416697,60.0)
    (2.6210896111,60.0)
    (2.6210896111,66.0)
};

\addplot[
    only marks,
    mark=*,
    mark size=2.5pt,
    color=blue
] coordinates {
    (1.3845782328,52.0)
    (1.6707399416,56.0)
    (2.3037416697,60.0)
    (2.6210896111,66.0)
};

\node[above,font=\footnotesize] at (axis cs:1.3845782328,52.0) {A};
\node[above,font=\footnotesize] at (axis cs:1.6707399416,56.0) {B};
\node[above,font=\footnotesize] at (axis cs:2.3037416697,60.0) {C};
\node[above,font=\footnotesize] at (axis cs:2.6210896111,66.0) {D};

\end{axis}
\end{tikzpicture}
\end{minipage}
\hfill
\begin{minipage}{0.38\linewidth}
\footnotesize
\begin{tabular}{@{}p{0.1\linewidth}p{0.7\linewidth}@{}}
A & Beam Search, population=1; constraints: \syntax + \comments \\
B & Beam Search, population=2; constraints: No Constraints \\
C & Sampling, population=3; constraints: \ut \\
D & Sequential Monte Carlo, population=5; constraints: \ut + \comments \\
\end{tabular}

\caption{\chadded{Pareto frontier for \codellama on the validation set. Gray points show all completed trials; blue points show the Pareto frontier in increasing time order (the blue line is a visual aid).} }
\label{fig:pareto_frontier}
\end{minipage}
\end{figure}




\paragraph{\textbf{Finding the best LM configuration}} 
Next, we run an optimisation experiment to choose the optimal LM configuration among \codellama, \deepseekshort and \mistralshort for accuracy and a soft time limit of 10 seconds, using the test/validation split of MBPP. 
This experiment also includes the \noexp constraint, which suppresses the generation of natural-language explanations.

Results are reported in Table~\ref{tab:public_private_results}, where the best emerging configuration is for \codellama. We also show 
the best configuration for each model. In summary, on MBPP, the best \codellama configuration outperforms \deepseekshort by 14\% and \mistral by 24\% indicating the importance of choosing the right LM for a given task. Notably, the best configuration is not the same for all models, for instance, \codellama does not require the \noexp constraint. In comparison to Table~\ref{tab:decoding_flat_public_private}, the best decoding strategy for \codellama remains the same; however, the constraint configuration slightly differs because of the larger space and the inherent non-determinism of sampling based algorithms.

\begin{table}
\centering
\scriptsize
\setlength{\tabcolsep}{4pt}
\begin{tabular}{@{}m{2cm} m{2.5cm} m{2.5cm} r r r r r@{}}
\toprule
\textbf{Language Model} &
\textbf{Decoding Function} &
\textbf{Constraint Config.} &
\makecell{\textbf{Pop.}\\\textbf{Size}} &
\makecell{\textbf{Validation}\\\textbf{Acc.(\%)}} &
\makecell{\textbf{Validation}\\\textbf{Time(s)}} &
\makecell{\textbf{Test}\\\textbf{Acc.(\%)}} &
\makecell{\textbf{Test}\\\textbf{Time(s)}} \\
\midrule
\codellama & Sequential Monte Carlo & \syntax\linebreak\ut & 5 & 66.00 & 2.67 & 61.27 & 3.29 \\
\deepseekshort     & Beam Search            & \noexp\linebreak\ut     & 5 & 52.00 & 9.47 & 50.66 & 9.24 \\
\mistralshort   & Beam Search            & \noexp\linebreak\ut   & 5 & 38.00 & 9.87 & 32.89 & 9.91 \\
\midrule
Baseline \codellama & Sampling & No Constraints & 1 & - & - & 36.1 & 1.72 \\
\bottomrule
\end{tabular}
\caption{Performance of each LM on validation (tuning) and test splits of the MBPP dataset.}
\label{tab:public_private_results}
\end{table}



\subsection{Comparison of \treecoder to Existing Frameworks}
\label{sec:comparison-to-existing-frameworks}

\paragraph{\textbf{Comparison to Hugging Face.}}
In Table~\ref{tab:runtime-HF}, we compare \treecoder against the Hugging Face \texttt{transformers} library on the MBPP benchmark using identical decoding settings. Despite its general-purpose abstractions, \treecoder achieves virtually identical accuracy---matching pass@5 and pass@1 scores across both beam search and sampling---while maintaining near-equivalent runtime. In particular, \treecoder's beam search is only about 25\% slower (\SI{2.8}{s} vs.\ \SI{2.1}{s}) and its sampling slightly faster. These results show that the framework's modularity and flexibility introduce negligible overhead compared to specialised implementations, while supporting far richer combinations of decoding strategies and constraints. 

\begin{table}[ht!]
\centering
\begin{tabular}{llrrr}
\toprule
\textbf{Name} & \textbf{Model} & \textbf{Time (s)} & \textbf{Pass@5} & \textbf{Pass@1}\\ 
\midrule
Beam Search (\treecoder) & \codellama & 2.79 & 61.0 & 51.0 \\ 
Beam Search (HF) & \codellama & 2.10 & 61.0 & 51.7 \\
\midrule
Sampling (\treecoder) & \codellama & 2.73 & 62.0 & 36.7 \\
Sampling (HF) & \codellama & 2.76 & 62.0 & 38.4 \\
\bottomrule
\end{tabular}
\caption{Runtime and accuracy comparison between \treecoder and Hugging Face Transformers on MBPP without constraints.}
\label{tab:runtime-HF}
\end{table}


\paragraph{\textbf{Comparison to IterGen.}}
We compare \treecoder with IterGen, which supports grammar-based backtracking but explores only a single decoding trajectory at a time. We re-implement IterGen's SQL decoding algorithm within \treecoder and reproduce its behaviour on Spider with 87.6\% identical queries and nearly identical accuracy (Table~\ref{tab:itergen vs tree}). 
\begin{chaddedblock}
    While \treecoder can instantiate IterGen's decoding algorithm, the reverse is not possible as IterGen
    supports only one single-trajectory decoding strategy that is tightly coupled with its constraints.
     In particular, IterGen cannot support the comparisons across decoding strategies in Sections~\ref{sec:targeted-comparison} and~\ref{sec:decoding-strategies}, nor the systematic exploration of decoding strategies and constraints in Section~\ref{sec:systematic-exploration}.
\end{chaddedblock}

\begin{table}[thbp]
\centering
\begin{tabular}{lllrr}
\toprule
\textbf{Name} & \textbf{Model} & \textbf{Constraints} & \textbf{Accuracy} & \makecell{\textbf{Identical}\\\textbf{Queries (\%)}} \\
\midrule
IterGen & Qwen2.5-Coder-1.5B & \syntax + \alias & 47.4 & 93.2 \\
IterGen (\treecoder) & Qwen2.5-Coder-1.5B & \syntax + \alias & 47.9 & 93.2 \\
\midrule
IterGen & Llama3-8B & \syntax + \alias & 36.4 & 90.4 \\
IterGen (\treecoder) & Llama3-8B & \syntax + \alias & 36.2 & 90.4 \\
\midrule
IterGen & Llama3.2-3B & \syntax + \alias & 23.3 & 79.1 \\
IterGen (\treecoder) & Llama3.2-3B & \syntax + \alias & 23.7 & 79.1 \\
\bottomrule
\end{tabular}
\caption{Comparison between IterGen's SQL decoding algorithm re-implemented in \treecoder{} and the original. The \syntax constraint enforces valid SQL syntax, while \alias ensures aliases refer to valid tables.}
\label{tab:itergen vs tree}
\end{table}

    \paragraph{\textbf{Comparison to iterative, post-hoc techniques.}}
An alternative to constrained decoding is iterative post-hoc validation, where the LM is repeatedly invoked until an output satisfies the constraints. Unlike \treecoder{}, which integrates constraint checking during decoding to guide and prune search, post-hoc methods evaluate constraints only after full generation. We consider two approaches: (i) \emph{generate-until-satisfying}, which samples until a valid solution is found, and (ii) \emph{prompt-refinement}, which augments the prompt with failed attempts. Both use Hugging Face sampling for CodeLlama-7B; results are in Table~\ref{fig:iterative-techniques}.

Unlike sequential sampling, \treecoder{} uses \emph{batched sampling} to generate candidates in parallel, achieving higher accuracy and lower runtime. Despite more node expansions, it is faster due to incremental \syntax and \comments checking, which prunes invalid sequences early, and GPU parallelism. Prompt-refinement performs worse, as the model often makes incremental edits rather than fully regenerating programs.

\begin{table}[htbp]
\begin{tabular}{lrlllrr}
\toprule
\textbf{Model} & \makecell{\textbf{Max Retries/}\\\textbf{Search Width}} & \textbf{Decoding} & \textbf{Constraints} & \textbf{Accuracy (\%)} & \makecell{\textbf{Expansions}} & \makecell{\textbf{Mean time}\\\textbf{taken (s)}} \\
\midrule
\codellama & 5 & (HF) Sampling \&  & All & 47.5 & 276 & 11.0 \\
 &  & prompt-refinement &  &  &  &  \\
\codellama & 5 & (HF) Sampling \& & All & 57.6 & 162 & 5.7 \\
& & generate-until-satisfying &  & & &  \\
\midrule
\codellama & 5 & (TreeCoder) Batched sampling & All & 60.7 & 316 & 3.23 \\
\end{tabular}
    \caption{\chadded{Comparison between \treecoder{} and iterative, post-hoc techniques. “All” combines \syntax, \ut and \comments constraints.}}
    \label{fig:iterative-techniques}
\end{table}

\subsection{Generalising to More Complex Tasks}
\subsubsection{LiveCodeBench~\cite{DBLP:conf/iclr/JainHGLYZWSSS25}}
LiveCodeBench is generally harder than MBPP: whereas MBPP was designed as a set of short programming tasks solvable by entry-level programmers, LiveCodeBench is constructed from contest-style problems collected from platforms such as LeetCode, AtCoder, and Codeforces, yielding a substantially more demanding evaluation setting.
As shown in Table~\ref{tab:live_code_bench}, adding the unit-test constraint (with a maximum runtime constraint of 30\,s) improves accuracy on LiveCodeBench by 41\% (from 22.8\% to 32.2\%). This shows that even on this more challenging benchmark, decoding-time semantic constraints can meaningfully guide the search toward more correct programs, albeit at the cost of roughly doubling the mean runtime.

\begin{table}[htbp]
\centering
\begin{tabular}{lrlllrr}
\toprule
\textbf{Model} & \textbf{Width} & \textbf{Decoding} & \textbf{Constraints} & \textbf{Accuracy (\%)} & \makecell{\textbf{Expansions}} & \makecell{\textbf{Mean time}\\\textbf{taken (s)}} \\
\midrule
Qwen2.5-Coder-7B-Instruct & 5 & Beam Search & - & 22.8 & 120 & 11.4 \\
Qwen2.5-Coder-7B-Instruct & 5 & Beam Search & Unit Tests & 32.2 & 192 & 22.4 \\
\end{tabular}
\caption{\chadded{Qwen2.5 Coder evaluation on LiveCodeBench. Maximum time per problem was set to 30 seconds. 
}}\label{tab:live_code_bench}
\end{table}

\subsubsection{Rust code generation}
To evaluate whether \treecoder{} extends beyond Python and SQL, we conduct an experiment on Rust code generation using a translated version of the MBPP dataset~\cite{orlanski_bc_mbpp}. Rust provides a useful stress test for constraint-based decoding because its typing and ownership rules make partial program validation more challenging than in dynamically typed languages.

\paragraph{\textbf{Constraint based on the Rust analyzer.}}
To enforce syntactic correctness during decoding, we integrate the \texttt{rust-analyzer} language server as an external constraint, which provides feedback on syntax and type errors. However, it does not provide detailed diagnostics for incomplete functions, which frequently arise during token-by-token generation.
To address this, we extract the partial parse tree from \texttt{rust-analyzer} and automatically close open brackets before checking the program.
Since this produces functions with incomplete return expressions, we append a \texttt{todo!()} statement at the end to ensure the return type remains valid.

\paragraph{\textbf{Decoding strategy.}} 
For Rust we use a Best-First Search decoding strategy. When a completion violates the Rust constraint, the search backtracks to the last syntactically valid line and continues from that point. 
Since \texttt{rust-analyzer} takes roughly 1\,s per check, we apply \emph{lazy constraint checking}, invoking it only on the highest-scoring node in the search frontier instead of all candidates.
As shown in Table~\ref{table-rust}, the Rust-analyzer constraint substantially increases the proportion of compilable programs, raising the compilation rate by 38\% (63.8\% to 88.3\%) showing that \treecoder{} can incorporate language-specific tooling and constraints beyond grammar-based settings.

\begin{table}[htbp]
\centering
\begin{tabular}{lrllllrr}
\toprule
\textbf{Model} & \textbf{Width} & \textbf{Decoding} & \textbf{Constraints} & \textbf{Compilation Rate (\%)} & \makecell{\textbf{Expansions}} & \makecell{\textbf{Mean time}\\\textbf{taken (s)}} \\
\midrule
\codellama & 10 & Best-First-Search & - & 63.8 & 103 & 4.7 \\
\codellama & 10 & Best-First-Search & Rust-Analyzer & 88.3 & 155 & 21.4 \\
\end{tabular}
\caption{\chadded{Results on the translated Rust MBPP dataset.}}
\label{table-rust}
\end{table}

%% file: sections/related.tex


Decoding determines how an LLM explores its output space when generating text or code.
While many studies introduce constrained decoding methods incorporating lexical, syntactic, or semantic rules~\cite{hokamp-2017, scholak-etal-2021-picard, lu-etal-2021-neurologic, banerjee2025crane, poesia-etal-2022-synchromesh}, most build on a small set of core search paradigms: sampling~\cite{alphaCode, chen2021evaluatinglargelanguagemodels, zhu2023hotcoldadaptivetemperature}, beam search~\cite{meister2021beamsearchanswerquestion, huang2018finishoptimalbeamsearch, he-etal-2023-empirical, hokamp-2017, scholak-etal-2021-picard, lu-etal-2021-neurologic}, SMC~\cite{SequentialMonteCarloMethods, sequentialMonteCarloApplications, loula2025syntacticsemanticcontrollarge}, and MCTS~\cite{guan2025rstarmathsmallllmsmaster, puri2025rolloutrouletteprobabilisticinference, lu2025lsrmctsalleviatinglongrange}.
Various works extend these algorithms by introducing constraints during generation. Such constraints are especially valuable in code generation, where violations lead to compilation or execution errors, and can significantly improve performance when integrated into search~\cite{princis2025enhancing}.
Constraint-aware beam search~\cite{hokamp-2017, lu-etal-2021-neurologic, banerjee2025crane} integrates lexical, logical, or grammatical restrictions into beam expansion, while systems such as Synchromesh~\cite{poesia-etal-2022-synchromesh} adapt sampling to enforce semantic or structural validity.
Despite their diversity, these approaches instantiate specific combinations of search policy and constraint enforcement—precisely the coupling our framework aims to decouple and generalise.
Next, we focus on grammar-constrained and grammar-aligned decoding, two major paradigms for ensuring structural and semantic validity during generation.

\emph{\textbf{Grammar-Constrained Decoding.}}
Grammar-constrained decoding (GCD) aims to generate sentences that conform to a formal grammar, making it applicable to tasks such as entity disambiguation, closed information extraction, constituency parsing, conflict resolution, and code generation~\cite{geng2024grammarconstraineddecodingstructurednlp, scholak2021picardparsingincrementallyconstrained}. 
GCD methods typically use a parser to mask tokens that would lead to invalid completions~\cite{geng2024grammarconstraineddecodingstructurednlp, scholak2021picardparsingincrementallyconstrained, poesia2022synchromeshreliablecodegeneration}, thereby ensuring syntactic well-formedness of the generated text.

\emph{\textbf{Grammar-Aligned Decoding.}}
While GCD enforces syntactic validity, it does not ensure semantic alignment with the underlying grammar. It may assign high probability to prefixes whose valid continuations are semantically implausible, distorting the output distribution~\cite{park2024grammaraligneddecoding}. 
To address this, grammar-aligned decoding (GAD) methods refine decoding using grammar feedback. 
ASAp~\cite{park2024grammaraligneddecoding} sequentially samples under grammar constraints and adjusts token probabilities; SMC~\cite{loula2025syntacticsemanticcontrollarge} maintains a population of constrained hypotheses that are periodically resampled; and MCMC~\cite{gonzalez2025constrainedsamplinglanguagemodelsshouldbeeasy} iteratively refines full completions via prefix-based edits. 
All three approaches converge to the true constrained distribution but differ in their trade-offs: MCMC can converge quickly on structured tasks, SMC benefits from GPU-friendly parallelisation at the cost of higher memory use, and ASAp and MCMC provide adaptive runtime budgets through sequential refinement.

\emph{\textbf{Constrained Decoding Frameworks.}}
The most closely related frameworks are \textit{SynCode}~\cite{DBLP:journals/corr/abs-2403-01632} and \textit{IterGen}~\cite{ugare2025itergeniterativesemanticawarestructured}. 
%
%
\chadded{
SynCode constrains LLM outputs using grammar rules, while IterGen extends it with semantic constraints. However, IterGen has key limitations: (i) it supports only a single decoding strategy tightly coupled with constraints, and (ii) it is inherently single-trajectory, lacking support for batched multi-sequence generation. As a result, it cannot implement population-based algorithms (e.g.\ beam search, SMC, sampling). \treecoder{} addresses these limitations through a unifying abstraction that decouples decoding and constraints, enabling both trajectory- and population-based generation, modular composition, and systematic optimisation.
}

%% file: sections/conclusions.tex
We introduced \treecoder, a modular framework for LLM constraint decoding that unifies diverse decoding strategies and constraint types within a single tree-based abstraction. Its flexible design enables automatic optimisation over decoding strategies, constraints and hyperparameters, allowing users to systematically explore and compare configurations through a shared interface. Experiments for Python, SQL and Rust show that enforcing constraints during decoding and jointly optimising decoding configurations with \treecoder significantly improves code-generation accuracy across models such as CodeLlama, Mistral, DeepSeek and Qwen.


%% file: sections/appendix.tex
\newpage
\label{sec:Appendix}

\section{\textbf{Comparison To Itergen}}
\label{sec:itergen}
IterGen allows traversing the search tree using grammatical units instead of tokens. This is done by incrementally parsing the currently generated sequence while keeping track of the grammatical role each token plays as well as the character start and end positions.

In figure \ref{fig:comparisonToIterger} we see a comparison between our and original implementation of the SQL backtracking algorithm proposed within the Itergen paper. The algorithm uses a best-first style greedy search to generate a grammatically valid continuation until a table or column name (lines 4-7). After a column name is generated, the algorithm verifies that the most recent column name exists (lines 8-9) and backtracks to the beginning of the column name (line 10) if it does not exist. To avoid infinite loops, the process allows for a fixed number of times that the algorithm can backtrack. Each time the algorithm applies the \textit{backward} function, the token scores that lead to the invalid completion are down-weighted. This makes those tokens less likely to be chosen again after backtracking.

We implemented the checks for the table and column names as part of the the main decoding loop instead of as a separate constraint functions. This is because the amount we want to backtrack varies depending on the type of constraint violation we have. For example, if a column violation is detected then we want to backtrack to the last column name. The tight coupling between constraints and using that information to decide how far to backtrack allows this algorithm to be very efficient at the cost of re-usability. 

Both implementations are logically equivalent; however, they differ in style. In our implementation, the SQL grammar is a constraint given to the expand\_nodes function. This modularity allows us to recognize some non-CFG languages using a combination of context free parsers.

In Itergen's implementation the forward function always uses a grammar to perform GCD until a grammatical unit is generated. The forward function also abstracts the next node's score calculation. This differs from our implementation where the score calculation is done within the main decoding function. 

Overall, each framework has its own strengths and weaknesses. The itergen framework treats grammatical units as first-class citizens and is therefore tailored for implementing constraints that require navigating the underlying grammar. Forward, backward and view functions all operate on grammatical units and enable easy checking for grammatical constraints. Our framework places more focus on constraint and decoding function modularity which allows us to support a wider range of constraints and decoding strategies at the cost of not specialising for GCD. In our framework, the first order citizen is a node and backward and view functions are performed by inspecting the node's state.

\begin{figure}
    \centering
    \begin{minipage}{0.95\linewidth}
        \begin{algorithm}[H]
        \begin{algorithmic}
            \Require $Tree$: The current search tree  
            \Require $prompt$: Prompt for the current problem.
            \Require LM: Language model
            \Require $\phi_{SQL}$: SQL grammar constraints
        \end{algorithmic}
        \end{algorithm}
    \end{minipage}
    \vspace{0.5em}
    \begin{subfigure}[t]{0.48\linewidth}
        \begin{algorithmic}[1]
            \State $node \gets Tree.root\_node$
            \State \color{gray} $schema \gets parse\_schema(prompt)$
            \State $iters = 0$ \color{black}
            \While {not $node.is\_complete$}
                \State $nodes \gets Expand(node, $LM$, \phi_{SQL})$ 
                \State $nodes \gets score\_probability(nodes)$
                \State $node \gets nodes[0]$
                \State $cols \gets view(child, `columns')$ \color{gray}
                \If {$cols[-1] \not \in schema$ and $iters < max\_iter$}
                    \State \color{black} $node \gets backward(cols[-1])$
                    \State \color{gray} $iters += 1$
                    \State \textbf{continue}
                \EndIf
                \State \color{black} $tables \gets view(child, `tables')$ \color{gray}
                \If {$tables[-1] \not \in schema$ and $iters < max\_iter$}
                    \State \color{black} $node \gets backward(tables[-1])$
                    \State \color{gray} $iters += 1$
                    \State \color{gray} \textbf{continue}
                \EndIf
            \EndWhile
            \State \color{black} \Return $Tree$
        \end{algorithmic}
        \subcaption{SQL Itergen decoding function (our implementation)}
    \end{subfigure}
    \hfill
    \begin{subfigure}[t]{0.48\linewidth}
        \begin{algorithmic}[1]
            \State $it.start(prompt, `SQL', $LM$)$
            \State \color{gray} $schema \gets parse\_schema(prompt)$
            \State $iters = 0$ \color{black}
            \While{not $it.finished()$}
                \State $out \gets it.forward('columns','tables')$
                \State
                \State
                \State $cols \gets it.view('columns')$ \color{gray}
                \If {$cols[-1] \not \in schema$ and $iters < max\_iter$}
                    \State \color{black} $it.backward('columns')$
                    \State \color{gray} $iters += 1$
                    \State \textbf{continue}
                \EndIf
                \State \color{black}$tables \gets it.view(`tables')$\color{gray}
                \If {$tables[-1] \not \in schema$ and $iters < max\_iter$}
                    \State \color{black} $it.backward('tables')$
                    \State \color{gray} $iters += 1$
                    \State \textbf{continue}
                \EndIf
            \EndWhile
            \State \color{gray}\Return $out$
        \end{algorithmic}
        \subcaption{SQL Itergen (original implementation)}
    \end{subfigure}

    \caption{Comparison between SQL \texttt{itergen} implementations. All identical lines are grayed out.}
    \label{fig:comparisonToIterger}
\end{figure}

\section{\textbf{Comparison To Hugging Face}}
\label{sec:hugging face}
Table~\ref{tab:runtime} shows that \treecoder achieves accuracy identical to Hugging Face's transformers library while maintaining near-equivalent runtime. Specifically, \treecoder’s beam search is only 25\% slower (2.79 s vs. 2.10 s) and its sampling slightly faster (2.73 s vs. 2.76 s) on the MBPP benchmark, with matching pass@5 scores (0.61–0.62). This demonstrates that the framework's general-purpose abstractions introduce minimal overhead compared to a specialised implementation, while providing far greater flexibility and extensibility across decoding strategies and constraint configurations.

\begin{table}[ht!]
\centering
\begin{tabular}{llrrr}
\toprule
\textbf{Name} & \textbf{Model} & \textbf{Time (s)} & \textbf{Pass@5} & \textbf{Pass@1}\\ 
\midrule
Beam Search (\treecoder) & \codellama & 2.79 & 61.0 & 51.0 \\ 
Beam Search (HF) & \codellama & 2.10 & 61.0 & 51.7 \\
\midrule
Sampling (\treecoder) & \codellama & 2.73 & 62.0 & 36.7 \\
Sampling (HF) & \codellama & 2.76 & 62.0 & 38.4 \\
\bottomrule
\end{tabular}
\caption{Runtime and accuracy comparison between \treecoder and Hugging Face Transformers on MBPP without constraints.}
\label{tab:runtime}
\end{table}

\section{Additional Experimental Evaluation}

\subsection{Optimising for Top-5 Accuracy and Time}
In Section~\ref{sec:systematic-exploration}, we identified optimal configurations for top-1 accuracy. We next examine how accuracy trades off against runtime efficiency when considering 5 generated solutions. Specifically, we search for decoding configurations that maximise top-5 accuracy while maintaining low average inference time per sample.

\begin{table}[h]
\centering
\caption{Optimising for top-5 accuracy (\%) and time taken in seconds. The objective value is calculated by subtracting time taken from top-5 accuracy.}
\begin{tabular}{llll}
\toprule
\makecell{\textbf{Decoding}\\\textbf{Function}} & 
\makecell{\textbf{Objective Value}} & 
\makecell{\textbf{Potential}\\\textbf{Function}} & 
\makecell{\textbf{Population}\\\textbf{Size}} \\
\midrule
Sampling & 64.26 & (none) & 5 \\
Sequential Monte Carlo & 63.94 & Python Syntax & 5 \\
Beam Search & 63.84 & (none) & 5 \\
Monte Carlo Tree Search & 57.26 & (none) & 5 \\
ASAP & 55.86 & Comments & 3 \\
\bottomrule
\end{tabular}
\label{tab:Validation-Across-Search}
\end{table}

The results in Table~\ref{tab:Validation-Across-Search} show that lightweight, population-based methods such as Sampling and SMC achieved the best trade-off between accuracy and runtime, whereas more exploratory algorithms like MCTS and ASAp incur higher runtimes with diminishing returns. Interestingly, the optimiser often preferred unconstrained or lightly constrained decoding (e.g. Python syntax only), suggesting that strict constraints can slow decoding. 

\subsection{Adaptability of \treecoder}
\label{sec:Treecoder-applications}

We now turn to \treecoder's broader adaptability, showing how the same framework can  enable systematic studies of inference-time scaling. In particular, we investigate how increasing computational resources at inference affects accuracy. This relationship is described by inference-time scaling laws. In particular, we ask if \textit{inference time scaling laws can be observed by changing the decoding function?} and secondly we ask \textit{what the effect of adding constraints is to inference time scaling?} 

Figure \ref{fig:scaling_laws_algorithm}, shows how many node expansions are needed for each algorithm to find 5 complete sequences when all constraints are used. We observe an increase in pass@1 accuracy when the number of node expansions increases. This is the expected result and shows that no single decoding function is better than another, meaning that the choice of the decoding function comes down to one's computational budget. In summary, we can clearly observe inference scaling laws by varying the computational budget allocated to scaling.

Figure \ref{fig:scaling_laws_constraints} shows how scaling laws differ when in the presence of constraints. It shows that when no constraints are used, beamsearch quickly hits a plateau in terms of pass@1 metric. While the pass@5 metric increases as when number of node expansions is increased, this is explained by the fact that larger beam sizes get more attempts at a solution being correct (e.g. beamsize of 4 will produce 2 more solutions than beamsize of 2). 

Interestingly, we see that introduction of constraints allows us to achieve much higher scaling before the accuracy levels off. This is because in highly constrained scenarios once a constraint violation has been detected in one branch, there may not be any valid prefixes in other branches that beamsearch could continue off of. This is easy to see when the beamsize is one and a constraint violation is detected, there are no other branches to switch to and thus the accuracy is equivalent. 


\begin{figure}[ht!]
\centering
\begin{subfigure}[t]{0.48\textwidth}
    \centering
    \begin{tikzpicture}
        \begin{axis}[
            width=\linewidth,
            height=7cm,
            xlabel={Number of Expansions},
            ylabel={Pass@1 Accuracy (\%)},
            xmin=100, xmax=1300,
            ymin=58, ymax=72,
            grid=major,
            grid style={dashed,gray!30},
            scatter/classes={
                Sampling={mark=*,blue},
                SMC={mark=*,red},
                BeamSearch={mark=*,green!60!black},
                MCTS={mark=*,purple},
                ASAP={mark=*,orange}
            },
            legend style={at={(0.5,-0.15)}, anchor=north, legend columns=3}
        ]

        \addplot[scatter,only marks,point meta=explicit symbolic]
        coordinates {
            (316,60.7) [Sampling]
            (332,61.4) [SMC]
            (706,63.0) [BeamSearch]
            (1099,65.3) [MCTS]
            (1158,68.9) [ASAP]
        };

        \node at (axis cs:316,60.7) [anchor=north] {Sampling};
        \node at (axis cs:332,61.4) [anchor=south] {SMC};
        \node at (axis cs:706,63.0) [anchor=north] {Beam Search};
        \node at (axis cs:1099,65.3) [anchor=north] {MCTS};
        \node at (axis cs:1158,68.9) [anchor=north] {ASAP};

        \end{axis}    
    \end{tikzpicture}
    \caption{Effect of changing the constraint function when all decoding functions stop after 5 complete valid nodes are found. 
    }
    \label{fig:scaling_laws_algorithm}
\end{subfigure}
\hfill
\begin{subfigure}[t]{0.48\textwidth}
    \centering
    \begin{tikzpicture}
    \begin{axis}[
        xlabel={Number of Expansions},
        ylabel={Pass@k Accuracy (\%)},
        legend style={at={(0.5,-0.2)},anchor=north,legend columns=2},
        width=\linewidth,
        height=7cm,
        grid=major,
        grid style={dashed,gray!30},
    ]

    \addplot+[mark=*, thick, blue]
        table[row sep=crcr] {
    46.8   49.41451991 \\
    98.8   50.58548009 \\
    203.0  50.58548009 \\
    400.5  50.35128806 \\
    };

    \addplot+[mark=square*, thick, blue, mark options={fill=blue}]
        table[row sep=crcr] {
    46.8   49.41451991 \\
    98.8   55.03512881 \\
    203.0  58.54800937 \\
    400.5  62.06088993 \\
    };

    \addplot+[mark=*, thick, red, mark options={fill=red}]
        table[row sep=crcr] {
    42.9   48.71194379 \\
    287.5  55.50351288 \\
    570.6  62.06088993 \\
    1156.6 66.04215457 \\
    };

    \addplot+[mark=square*, thick, red]
        table[row sep=crcr] {
    42.9   48.71194379 \\
    287.5  56.90866511 \\
    570.6  65.10538642 \\
    1156.6 70.25761124 \\
    };

    \end{axis}
    \end{tikzpicture}
    \caption{Effect of varying beam size (1,2,4,8) on pass@1 and pass@5 accuracy. The X-axis shows the number of expansions required to find the corresponding number of complete sequences. Red lines indicate constrained decoding; blue lines indicate unconstrained decoding.}
    \label{fig:scaling_laws_constraints}
\end{subfigure}

\caption{Comparison of decoding performance under different scaling and constraint conditions.}
\label{fig:combined_scaling_laws}
\end{figure}

\subsection{Lines of Code}
\begin{table}[ht!]
\centering
\begin{tabular}{l l r}
\toprule
\textbf{Function Type} & \textbf{Function Name} & \textbf{LoC} \\
\midrule
Decoding Function & Itergen & 130 \\
Decoding Function & MCTS & 113 \\
Decoding Function & ASAP & 70 \\
Decoding Function & SMC & 53 \\
Decoding Function & Sampling & 39 \\
Decoding Function & Beam Search & 39 \\
Constraint Function & \syntax & 26 \\
Constraint Function & \ut & 26 \\
Constraint Function & \comments & 7 \\
Constraint Function & SQL Syntax & 32 \\
Constraint Function & \noexp & 13 \\
Constraint Function & \sig & 49 \\
Constraint Function & \type & 49 \\
Aggregation Function & Default Aggregation & 58 \\
& Function &  \\
Termination Condition & Default Termination & 26 \\
 & Condition &  \\
\bottomrule
\end{tabular}
\caption{The lines of code count was obtained by excluding empty lines and comments using the cloc utility. Line length does not exceed 88 characters.}
\label{tab:function_summary}
\end{table}